%% file: main.tex
\documentclass[final]{cvpr}
\usepackage{times}
\usepackage{amsmath}
\usepackage{amssymb}
\usepackage{cuted}
\usepackage{graphicx}
\usepackage{xcolor}
\usepackage{soul}
\usepackage{subcaption}
\usepackage{booktabs}
\usepackage{arydshln}
\usepackage[pagebackref=true,breaklinks=true,colorlinks,bookmarks=false]{hyperref}
\usepackage[numbers,sort]{natbib} 
\usepackage{cleveref}
\usepackage{comment}
\usepackage{algorithm}
\usepackage{listings}
\usepackage[toc,page]{appendix}

\def\cvprPaperID{1283} 
\def\confYear{CVPR 2021}

\newcommand{\dataset}[0]{{VGG-SS}}
\newcommand\algcomment[1]{\def\@algcomment{\footnotesize#1}}

\makeatletter
\renewcommand{\paragraph}{%
  \@startsection{paragraph}{4}%
  {\z@}{0.5em}{-1em}%
  {\normalfont\normalsize\bfseries}%
}
\makeatother

\title{Localizing Visual Sounds the Hard Way}
\author{Honglie Chen, Weidi Xie, Triantafyllos Afouras, Arsha Nagrani \\[3pt]
Andrea Vedaldi, Andrew Zisserman\\[2pt]
{\tt\small \{hchen, weidi, afourast, arsha, vedaldi, az\}@robots.ox.ac.uk}\\ [3pt]
VGG, Department of Engineering Science, University of Oxford, UK\\[3pt]
\url{http://www.robots.ox.ac.uk/~vgg/research/lvs/} \vspace{-10pt}}

\begin{document}
\vspace{-15pt}
\maketitle
\pagestyle{empty}
\thispagestyle{empty}

\input{fig_table/Fig-teaser}
\begin{abstract}
\vspace{-5pt}
The objective of this work is to localize sound sources that are visible in a video without using manual annotations.
Our key technical contribution is to show that, by training the network to explicitly discriminate challenging image fragments, even for images that do contain the object emitting the sound, we can significantly boost the localization performance.
We do so elegantly by introducing a mechanism to mine hard samples and add them to a contrastive learning formulation automatically.
We show that our algorithm achieves state-of-the-art performance on the popular Flickr SoundNet dataset.
Furthermore, we introduce the VGG-Sound Source (VGG-SS) benchmark, a new set of annotations for the recently-introduced VGG-Sound dataset, where the sound sources visible in each video clip are explicitly marked with bounding box annotations.
This dataset is 20 times larger than analogous existing ones, contains 5K videos spanning over 200 categories, and, differently from Flickr SoundNet, is video-based.
On VGG-SS, we also show that our algorithm achieves state-of-the-art performance against several baselines. 
\end{abstract}
\input{section/introduction}
\input{section/related_work}

\input{section/method}
\input{section/dataset}
\input{section/experiments}
\input{section/results}
\input{section/conclusion}

\section*{Acknowledgements}
This work is supported by the UK EPSRC CDT in Autonomous Intelligent Machines and Systems, the
Oxford-Google DeepMind Graduate Scholarship, the Google PhD Fellowship, and EPSRC Programme Grants Seebibyte EP/M013774/1 and VisualAI EP/T028572/1.
{\small\bibliographystyle{ieee_fullname}\bibliography{shortstrings,vgg_local,vgg_other,refs,vedaldi_specific,vedaldi_general}}

\clearpage
\onecolumn
\input{section/appendix}
\end{document}

%% file: fig_table/Fig-teaser.tex
\begin{strip}
\centering
\includegraphics[width=.99\textwidth]{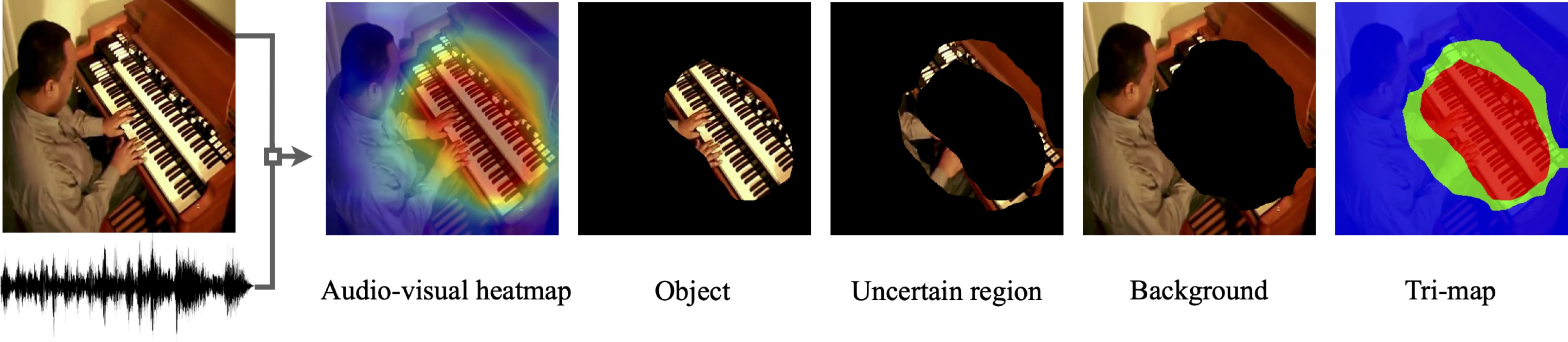}
\vspace{-5pt}
\captionof{figure}{\textbf{Visual Sound Source Localisation:} We localise sound sources in videos without manual annotation. Our key contribution is an automatic negative mining technique through differentiable thresholding of a cross-modal correspondence score map, 
the background regions with low correlation to the given sound as `hard negatives',
and the regions in the Tri-map is`ignored' in a contrastive learning framework.}
\label{fig:teaser}
\end{strip}

%% file: section/introduction.tex
\section{Introduction}\label{sec:intro}

While research in computer vision largely focuses on the visual aspects of perception, natural objects are characterized by much more than just appearance.
Most objects, in particular, emit sounds, either in their own right, or in their interaction with the environment --- think of the bark of a dog, or the characteristic sound of a hammer striking a nail. 
A full understanding of natural objects should not ignore their acoustic characteristics.
Instead, modelling appearance and acoustics jointly can often help us understand them better and more efficiently.
For example, several authors have shown that it is possible to use sound to discover and localize objects automatically in videos, 
without the use of any manual supervision~\cite{Arandjelovic18objects,senocak2018learning,harwath2018jointly,owens18audio-visual,Hu_2019_CVPR,Afouras20b}.

In this paper, we consider the problem of localizing `visual sounds', 
\ie~visual objects that emit characteristics sounds in videos.
Inspired by prior works~\cite{Arandjelovic18objects,senocak2018learning,harwath2018jointly}, 
we formulate this as finding the correlation between the visual and audio streams in videos.
These papers have shown that not only can this correlation be learned successfully, but that, once this is done, 
the resulting convolutional neural networks can be `dissected' to localize the sound source spatially, thus imputing it to a specific object.
However, other than in the design of the architecture itself, there is little in this prior work meant to improve the localization capabilities of the resulting models.
In particular, while several models~\cite{Arandjelovic18objects,Afouras20b,senocak2018learning} do incorporate a form of spatial attention which should also help to localize the sounding object as a byproduct, these may still fail to provide a good \emph{coverage} of the object, often detecting too little or too much of it.

In order to address this issue, we propose a new training scheme that explicitly seeks to spatially localize sounds in video frames. Similar to object detection~\cite{viola01brobust}, in most cases only a small region in the image contains an object of interest, 
in our case a `sounding' object, with the majority of the image often being `background' which is not linked to the sound.
Learning accurate object detectors involves explicitly seeking for these background regions, prioritizing those that could be easily confused for the object of interest, also called \emph{hard negatives}~\cite{viola01brobust,dalal05histogram,Girshick14,shrivastava2016training,Ren16,Lin2017}.
Given that we lack supervision for the location of the object making the sound, however, we are unable to tell which boxes are positive or negative.
Furthermore, since we seek to solve the localization rather than the detection problem, we do not even have bounding boxes to work with, as we seek instead a segmentation of the relevant image area.

In order to incorporate hard evidence in our unsupervised (or self-supervised) setting, 
we propose an automatic background mining technique through differentiable thresholding, \ie regions with low correlation to the given sound are incorporated into a negatives set for contrastive learning.
Instead of using hard boundaries, we note that some regions may be uncertain, and hence we introduce the concept of a Tri-map into the training procedure,
leaving an `ignore' zone for our model.
To our knowledge, 
this is the first time that background regions have been explicitly considered when solving the sound source localization problem.
We show that this simple change significantly boosts sound localization performance on standard benchmarks, 
such as Flickr SoundNet~\cite{senocak2018learning}.

To further assess sound localization algorithms, 
we also introduce a new benchmark, based on the recently-introduced VGG-Sound dataset~\cite{Chen20a},
where we provide high-quality bounding box annotations for `sounding' objects, 
\ie~objects that produce a sound, for more than 5K videos spanning 200 different categories.
This dataset is $20\times$ larger and more diverse than existing sound localization benchmarks, 
such as Flickr SoundNet (the latter is also based on still images rather than videos).
We believe this new benchmark, which we call VGG-Sound Source, or VGG-SS for short, 
will be useful for further research in this area.
In the experiments, we establish several baselines on this dataset, 
and further demonstrate the benefits of our new algorithm.

%% file: section/related_work.tex
\section{Related Work}\label{sec:related}

\subsection{Audio-Visual Sound Source Localization}



Learning to localize sound sources by exploiting the natural co-occurrence of
visual and audio cues in videos has a long history.
Early attempts to solve the task used shallow probabilistic models~\cite{hershey1999audio,fisher2000learning,kidron2005pixels}, or proposed
segmenting videos into spatio-temporal tubes and associating those to the audio signal through
canonical correlation analysis (CCA)~\cite{izadinia2012multimodal}.

Modern approaches solve the problem using deep neural networks --- typically employing a dual stream,
trained with a contrastive loss by exploiting the audio-visual correspondence,
\ie matching audio and visual representations extracted from the same video.
For example, 
{}\cite{Arandjelovic18objects, senocak2018learning,harwath2018jointly, Ramaswamy_2020_WACV} associate the appearance
of objects with their characteristic sounds or audio narrations;
{Hu~\etal~}\cite{Hu_2019_CVPR} first cluster audio and visual representations within each modality, 
followed by associating the resulting centroids with contrastive learning;
{Qian~\etal~}\cite{qian2020multiple} proposed a weakly supervised approach, 
where the approximate locations of the objects are obtained from CAMs to bootstrap the model training.
Apart from using correspondence,
Owens and Efros~\cite{Owens2018b} also localize sound sources through synchronization, 
a related objective also investigated in earlier works~\cite{Marcheret15, chung16}, while~\cite{khosravan2018attention} incorporate explicit attention in this model.
Afouras~\etal~\cite{Afouras20b} also exploit audio-visual concurrency to train a video model
that can distinguish and group instances of the same category.

Alternative approaches solve the task using an audio-visual source separation objective.
For example {Zhao~\etal~}\cite{zhao2018sound} employ a mix-and-separate approach to learn to
associate pixels in video frames with separated audio sources, while {Zhao~\etal~}\cite{zhao2019sound} extends this method by providing the model with motion information through optical flow.
{Rouditchenko~\etal~}\cite{rouditchenko2019self} train a two-stream model to co-segment video and audio, producing heatmaps that roughly highlight the object according to the audio semantics.
These methods rely on the availability of videos containing single-sound sources, usually found in well curated datasets.
In other related work, {Gan~\etal~}\cite{gan2019self} learn to detect cars from stereo sound, by distilling video object detectors, while {Gao~\etal~}~\cite{gao2019visual} lift mono sound to stereo by leveraging spatial information.


\subsection{Audio-Visual Localization Benchmarks}

Existing audio-visual localization benchmarks are summarised in \Cref{tab:stat_comp} (focusing on the test sets).
The Flickr SoundNet sound source localization benchmark~\cite{senocak2018learning} 
is an annotated collection of single frames randomly sampled from videos of the Flickr SoundNet dataset~\cite{aytar16soundnet,Thomee16}.
It is currently the standard benchmark for the sound source localization task;
we discuss its limitations in Section~\ref{sec:dataset}, where we introduce our new benchmark.
The Audio-Visual Event (AVE) dataset~\cite{tian2018audio}, contains 4,143 10 second video clips spanning 28 audio-visual event categories with temporal boundary annotations.
LLP~\cite{tian2020avvp} contains of 11,849 YouTube video clips spanning 25 categories for a total of 32.9 hours collected from AudioSet~\cite{Gemmeke17}.
The development set is sparsely annotated with object labels, while the test set contains dense video and audio sound event labels on the frame level.
Note that the AVE and LLP test sets contain only temporal localisation of sounds (at the frame level), with no spatial bounding box annotation.

\input{fig_table/Tab-testset_comp}

%% file: fig_table/Tab-testset_comp.tex
\begin{table}[!htb]
\vspace{1mm}
\begin{center}
\small
 \begin{tabular}{lcccc} 
 \toprule
 Benchmark Datasets & \# Data & \# Classes &  Video &  BBox\\ 
 \midrule
 Flickr SoundNet~\cite{senocak2018learning} & 250 & $\sim$50$\ddagger$  & $\times$ & $\checkmark$ \\ 

 AVE~\cite{tian2018audio}$\dagger$  & 402 &  28 &  $\checkmark$& $\times$\\

 LLP~\cite{tian2020avvp}$\dagger$ & 1,200 & 25  & $\checkmark$ & $\times$\\

 VGG-SS & 5,158 & $220$ & $\checkmark$ & $\checkmark$\\
 \bottomrule
\end{tabular}
\caption{ 
Comparison with the existing sound-source localisation benchmrks. Note that  VGG-SS has more images and classes. $\dagger$These datasets contain only temporal localisation of sounds, not spatial localisation. $\ddagger$ We determined this via manual inspection.  }
\label{tab:stat_comp}
\vspace{-5mm}
\end{center}
\end{table}

%% file: section/method.tex
\vspace{-5pt}
\section{Method}\label{sec:method}

\input{fig_table/Fig-model}

Our goal is to localize objects that make characteristic sounds in videos, without using any manual annotation.
Similar to prior work~\cite{Arandjelovic18objects}, we use a two-stream network to extract visual and audio representations from unlabelled video.
For localization, we compute the cosine similarity between the audio representation and the visual representations extracted convolutionally at different spatial locations in the images.
In this manner, we obtain a positive signal that pulls together sounds and relevant spatial locations.
For learning, we also need an opposite negative signal.
A weak one is obtained by correlating the sound to locations in other, likely irrelevant videos.
Compared to prior work~\cite{Arandjelovic18objects,Afouras20b}, 
our key contribution is to \emph{also} explicitly seek for hard negative locations that contain background or non-sounding objects in the \emph{same} images that contain the sounding ones, leading to more selective and thus precise localization. An overview of our architecture can be found in \Cref{Fig-model}.


While the idea of using hard negatives is intuitive, an effective implementation is less trivial.
In fact, while we seek for hard negatives, there is no hard evidence for whether any region is in fact positive (sounding) or negative (non-sounding) as videos are unlabelled.
An incorrect classification of a region as positive or negative can throw off the localization algorithm entirely.
We solve this problem by using a robust contrastive framework that combines soft thresholding and Tri-maps, 
which enables us to handle uncertain regions effectively.


In \cref{sec:rep,sec:corr,sec:contrast_oracle} we first describe the task of audio-visual localization using contrastive learning in its \emph{oracle} setting, assuming, for each visual-audio pair, 
we do have the ground-truth annotation for which region in the image is emitting the sound.
In \cref{sec:selfsup_loc}, we introduce our proposed idea, which replaces the \emph{oracle}, and discuss the difference between our method and existing approaches.

\subsection{Audio-Visual Feature Representation}\label{sec:rep}

Given a short video clip with $N$ visual frames and audio, and considering the center frame as visual input,
\ie~$X=\{I, a\}, I\in \mathbb{R}^{3\times H_v \times W_v}, 
a\in\mathbb{R}^{1\times H_a \times W_a}$.
Here, $I$ refers to the visual frame,
and $a$ to the spectrogram of the raw audio waveform.
In this manner, representations for both modalities can be computed by means of CNNs, 
which we denote respectively  $f(\cdot; \theta_1)$ and $g(\cdot; \theta_2)$.
For each video $X_i$, we obtain visual and audio representations:
\begin{align}
V_i &= f(I_i; \theta_1), \quad V_i\in\mathbb{R}^{c\times h\times w}, \label{e:videorep}
\\
A_i &= g(a_i; \theta_2), \quad A_i\in\mathbb{R}^{c}. \label{e:audorep}
\end{align}
Note that both visual and audio representation have the same number of channels $c$, which allows to compare them by using dot product or cosine similarity.
However, the video representation also has a spatial extent $h\times w$, which is essential for spatial localization.

\subsection{Audio-Visual Correspondence}\label{sec:corr}

Given the video and audio representations of~\cref{e:videorep,e:audorep}, 
we put in correspondence the audio of clip $i$ with the image of clip $j$ by computing the cosine similarity of the representations, using the audio as a probe vector:
\begin{align*}
[S_{i \rightarrow j}]_{uv} =
\frac
{\langle A_i, [V_j]_{:uv}\rangle}
{\|A_i\|~\|[V_i]_{:uv}\|},
\quad
uv \in [h] \times [w].
\end{align*}
This results in a map $S_{i \rightarrow j} \in \mathbb{R}^{h \times w}$ indicating how strongly each image location in clip $j$ responds to the audio in clip $i$. To compute the cosine similarity, the visual and audio features are $L^2$ normalized.
Note that we are often interested in correlating images and audio from the same clip, which is captured by setting $j=i$.

\subsection{Audio-Visual Localization with an Oracle}\label{sec:contrast_oracle}

In the literature, training models for audio-visual localization has been treated as learning the correspondence between these two signals, and formulated as contrastive learning~\cite{senocak2018learning,Arandjelovic18objects,Hu_2019_CVPR,Afouras20b,qian2020multiple}.

Here, before diving into the self-supervised approach, 
we first consider the \emph{oracle} setting for the contrastive learning 
where ground-truth annotations are available.
This means that we are given a training set
$
\mathcal{D} = \{d_1, d_2, \dots, d_k\},
$
where each training sample $d_i = (X_i, m_i)$ consists of a audio-visual sample $X_i$, as given above, plus a segmentation mask $m_i \in \mathbb{B}^{h\times w}$ with ones for those spatial locations that overlap with the object that emits the sounds, and zeros elsewhere.
During training, the goal is therefore to jointly optimize $f(\cdot;\theta_1)$ and $g(\cdot; \theta_2)$, such that $S_{i \rightarrow i}$ gives high responses only for the region that emits the sound present in the audio.
In this paper, we consider a specific type of contrastive learning, namely InfoNCE~\cite{oord2018representation,Han19}.

\paragraph{Optimization.}

For each clip $i$ in the dataset (or batch), we define the positive and negative responses as:
\begin{align*}
P_i &=
\frac{1}{|m_i|} 
\langle  m_i, ~S_{i \rightarrow i} \rangle,
\\
N_i &=
\underbrace{
\frac{1}{|\mathbf{1} - m_i|}
\langle \mathbf{1} - m_i,~ S_{i \rightarrow i} \rangle
}_{\text{hard negatives}}
+
\underbrace{
\frac{1}{hw} \sum_{i\not= j}
\langle \mathbf{1}, ~ S_{i\rightarrow j}\rangle
}_{\text{easy negatives}}.
\end{align*}
where $\langle \cdot, \cdot \rangle$ denotes Frobenius inner product.
To interpret this equation, 
note that the inner product simply sums over the element-wise product of the specified tensors and that
$\mathbf{1}$ denotes a $h\times w$ tensor of all ones.
The first term in the expression for $N_i$ refers to the \emph{hard negatives}, 
calculated from the ``background''~(regions that do not emit the characteristic sound) within the same image,
and the second term denotes the easy negatives, 
coming from other images in the dataset.
The optimization objective can therefore be defined as:
\begin{align*}
\centering
\mathcal{L} = 
-\frac{1}{k}\sum_{i=1}^k
\left[ \log \frac{\exp(P_i)}{\exp(P_i) + \exp(N_i)} \right]
\end{align*}

\paragraph{Discussion.}

Several existing approaches~\cite{senocak2018learning, harwath2018jointly, Arandjelovic18objects,Afouras20b} to self-supervised audio-visual localization are similar.
The key difference lies in the way of constructing the positive and negative sets.
For example, in~\cite{senocak2018learning} a heatmap generated by using the soft-max operator is used to pool the positives
and images from other video clips are treated as negatives;
instead, in~\cite{Arandjelovic18objects}, positives come from max pooling the correspondence map, 
$S_{i \rightarrow i}$ and the negatives from max pooling $S_{i \rightarrow j}$ for $j \neq i$.
Crucially, all such approaches have missed the \emph{hard negatives} term defined above,
computed from the background regions within the same images that do contain the sound.
Intuitively this term is important to obtain a shaper visual localization of the sound source;
however, while this is easy to implement in the oracle setting, obtaining hard negatives in self-supervised training requires some care, as discussed next.

\subsection{Self-supervised Audio-Visual Localization}\label{sec:selfsup_loc}

In this section, we describe a simple approach for replacing the oracle, 
and continuously bootstrapping the model to achieve better localization results.
At a high level, the proposed idea inherits the spirit of self-training,
where predictions are treated as pseudo-ground-truth for re-training.

Specifically, given a dataset $\mathcal{D} = \{X_1, X_2, \dots, X_k\}$ 
where only audio-visual pairs are available (but not the masks $m_i$),
the correspondence map $S_{i \rightarrow i}$ between audio and visual input can be computed in the same manner as \cref{sec:corr}.
To get the pseudo-ground-truth mask $\hat{m}_i$, 
we could simply threshold the map $S_{i\rightarrow i}$:
\[
\hat{m}_i= 
\begin{cases}
    1              ,& \text{if } S_{i \rightarrow i} \geq \epsilon \\
    0,              & \text{otherwise}
\end{cases}
\]
Clearly, however, this thresholding, which uses the Heaviside function,
is not differentiable.
Next, we address this issue by relaxing the thresholding operator.

\paragraph{Smoothing the Heaviside function.}

Here, we adopt a smoothed thresholding operator in order to maintain the end-to-end differentiability of the architecture:
\begin{align*}
\hat{m}_i = \text{sigmoid}((S_{i \rightarrow i}- \epsilon)/ \tau)
\end{align*}
where $\epsilon$ refers to the thresholding parameter, 
and $\tau$ denotes the temperature controlling the sharpness.

\paragraph{Handling uncertain regions.}

Unlike the oracle setting,
the pseudo-ground-truth obtained from the model prediction may potentially be noisy,
we therefore propose to set up an ``ignore'' zone between the positive and negative regions, allowing the model to self-tune.
In the image segmentation literature, 
this is often called a Tri-map and is also used for matting~\cite{Chuang02, tao2018scale}.
Conveniently,
this can be implemented by applying two different $\epsilon$'s, 
one controlling the threshold for the positive part and the other for the negative part of the Tri-map.

\paragraph{Training objective.}

We are now able to replace the oracle while computing the positives and negatives automatically.
This leads to our final formulation:
\begin{align*}
\hat{m}_{ip} &= \operatorname{sigmoid}((S_{i \rightarrow i} - \epsilon_p) / \tau)\\
\hat{m}_{in} &= \operatorname{sigmoid}((S_{i \rightarrow i} - \epsilon_n) / \tau)\\
P_i &= \frac{1}{|\hat{m}_{ip}|} \langle \hat{m}_{ip}, ~ S_{i \rightarrow i}\rangle \\
N_i &= 
\frac{1}{|\mathbf{1}- \hat{m}_{in}|} \langle \mathbf{1}- \hat{m}_{in}, ~ S_{i \rightarrow i} \rangle
+
\frac{1}{hw} \sum_{j \neq i} \langle \mathbf{1}, ~S_{i \rightarrow j} \rangle 
\\
\mathcal{L} &= 
-\frac{1}{k}\sum_{i=1}^k
\left[ \log \frac{\exp(P_i)}{\exp(P_i) + \exp(N_i)} \right]
\end{align*}
where $\epsilon_p$ and $\epsilon_n$ are two thresholding parameters~(validated in experiment section), 
with $\epsilon_p > \epsilon_n$.
For example if we set  $\epsilon_p = 0.6$ and $\epsilon_n = 0.4$, regions with correspondence scores above $0.6$ are considered positive and bellow $0.4$ negative, 
while the areas falling within the $[0.4,0.6]$ range are treated as ``uncertain'' regions 
and ignored during training (\Cref{Fig-model}).




%% file: fig_table/Fig-model.tex

\begin{figure*}[t]
\centering
\includegraphics[width=0.95\textwidth]{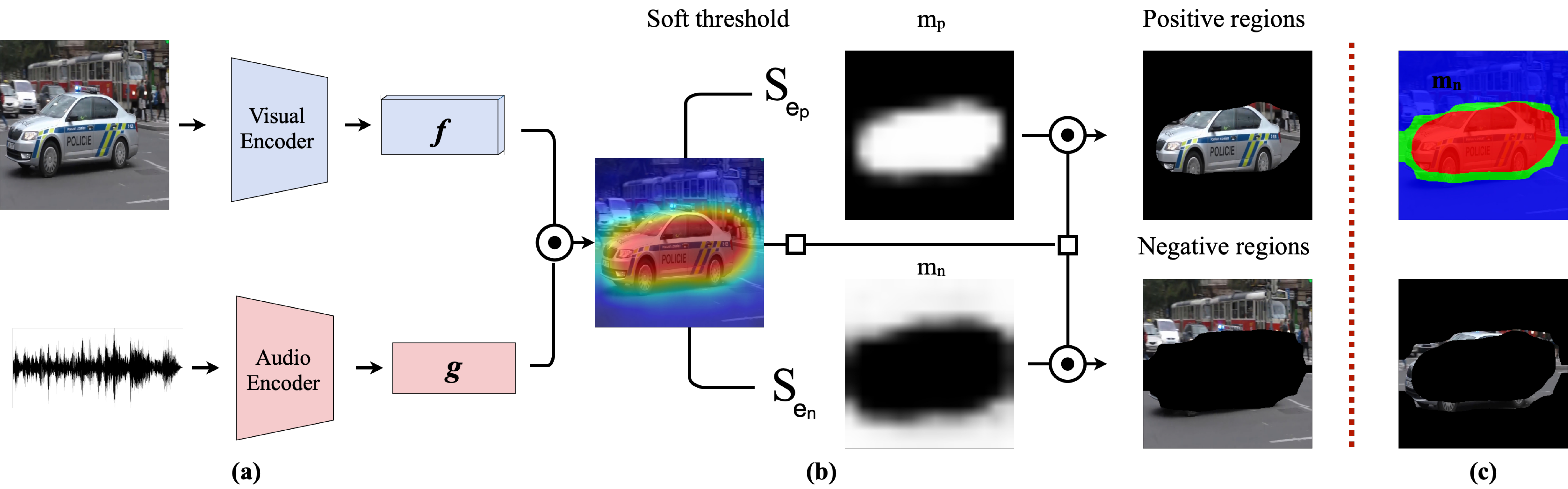}
\caption{\textbf{Architecture Overview}. 
We use an audio-visual pair as input to a dual-stream network shown in \textbf{(a)}, 
$f(\cdot;\theta_1)$ and $g(\cdot;\theta_2)$, denoting the visual and audio feature extractor respectively.
Cosine similarity between the audio vector and visual feature map is then computed, giving us a heatmap of size $14 \times 14$.
\textbf{(b)} demonstrates the soft threshold being applied twice with different parameters,
generating positive, negative regions. The final Tri-map and the uncertain region are highlighed in \textbf{(c)}.
}
\vspace{-5mm}
\label{Fig-model}
\end{figure*}

%% file: section/dataset.tex
\section{The VGG-Sound Source Benchmark}\label{sec:dataset}

As mentioned in Section~\ref{sec:related}, the SoundNet-Flickr sound source localization benchmark~\cite{senocak2018learning} is commonly used for evaluation in this task.
However, we found it to be unsatisfactory in the following aspects:
i) both the number of total instances (250) and sounding object categories (approximately 50) that it contains are limited,
ii) only certain reference frames are provided, instead of the whole video clip, which renders it unsuitable for the evaluation of video models, and 
iii) it provides no object category annotations.





In order to address these shortcomings, we build on the recent VGG-Sound dataset~\cite{Chen20a} and introduce \dataset, an audio-visual localization benchmark based on videos collected from YouTube. 

\subsection{Test Set Annotation Pipeline} 

In the following sections, we describe a semi-automatic procedure to annotate the objects that emit sounds with bounding boxes,
which we apply to obtain \dataset~with over 5k video clips, spanning 220 classes.

\paragraph{(1) Automatic bbox generation.}

We use the entire VGG-Sound test set, containing 15k 10-second video clips,
and extract the center frame from each clip.
We use a Faster R-CNN object detector~\cite{Ren16} pretrained on OpenImages to predict the bounding boxes of all relevant objects.
Following~\cite{Chen20a}, we use a word2vec model to match visual and audio categories that are semantically similar.
At this stage, there are roughly 8k frames annotated automatically.

\paragraph{(2) Manual image annotation.}

We then annotate the remaining frames manually. 
There are three main challenges at this point:
(i) there are cases where localization is extremely difficult or impossible, 
either because the object is not visible~(e.g. in extreme lighting conditions), 
too small (`mosquito buzzing'), or is diffused throughout the frame (`hail', `sea waves', `wind'); 
(ii) the sound may originate either from a single object, 
or from the interactions between multiple objects and a consistent annotation scheme must be decided upon; 
and finally (iii), there could be multiple instances of the same class in the same frame, 
and it is challenging to know which of the instances are making the sound from a single image. 

\input{fig_table/Fig-testset}

We address these issues in three ways:
First, we remove categories~(\eg~mainly environmental sounds such as wind, hail etc) 
that are challenging to localize, roughly 50 classes;
Second, as illustrated in Figure~\ref{fig:sub1}, 
when the sound comes from the interaction of multiple objects, we annotate a tight region surrounding the interaction point;
Third, if there are multiple instances of the same sounding object category in the frame, 
we annotate each separately when there are less than 5 instances and they are separable, otherwise a single bounding box is drawn over the entire region, as shown in the top left image (`human crowd') in \Cref{fig:sub1}.

\paragraph{(3) Manual video verification.}

Finally, we conduct manual verification on videos using the VIA software~\cite{Dutta19a}.
We do this by watching the 5-second video around every annotated frame, to ensure that the sound corresponds with the object in the bounding box.
This is particularly important for the cases where there are multiple candidate instances present in the frame, however, only one is making the sound, \eg~human singing.


The statistics after every stage of the process and the final dataset are summarised in \Cref{tab:stat_after_steps}.
The first stage generates bounding box candidates for the entire VGG-Sound test set (309 classes, 15k frames); the manual annotation process then removes unclear classes and frames, resulting in roughly 260 classes and 8k frames.
Our final video verification further cleans up the the test set, yielding a high-quality large-scale audio-visual benchmark --- VGG-Sound Source~(\dataset), 
which is 20 times larger than the existing one~\cite{senocak2018learning}.

\input{fig_table/Tab-testset_stat}


%% file: fig_table/Fig-testset.tex
\begin{figure*}
\centering
\begin{subfigure}{.49\textwidth}
  \centering
  \includegraphics[height=3cm]{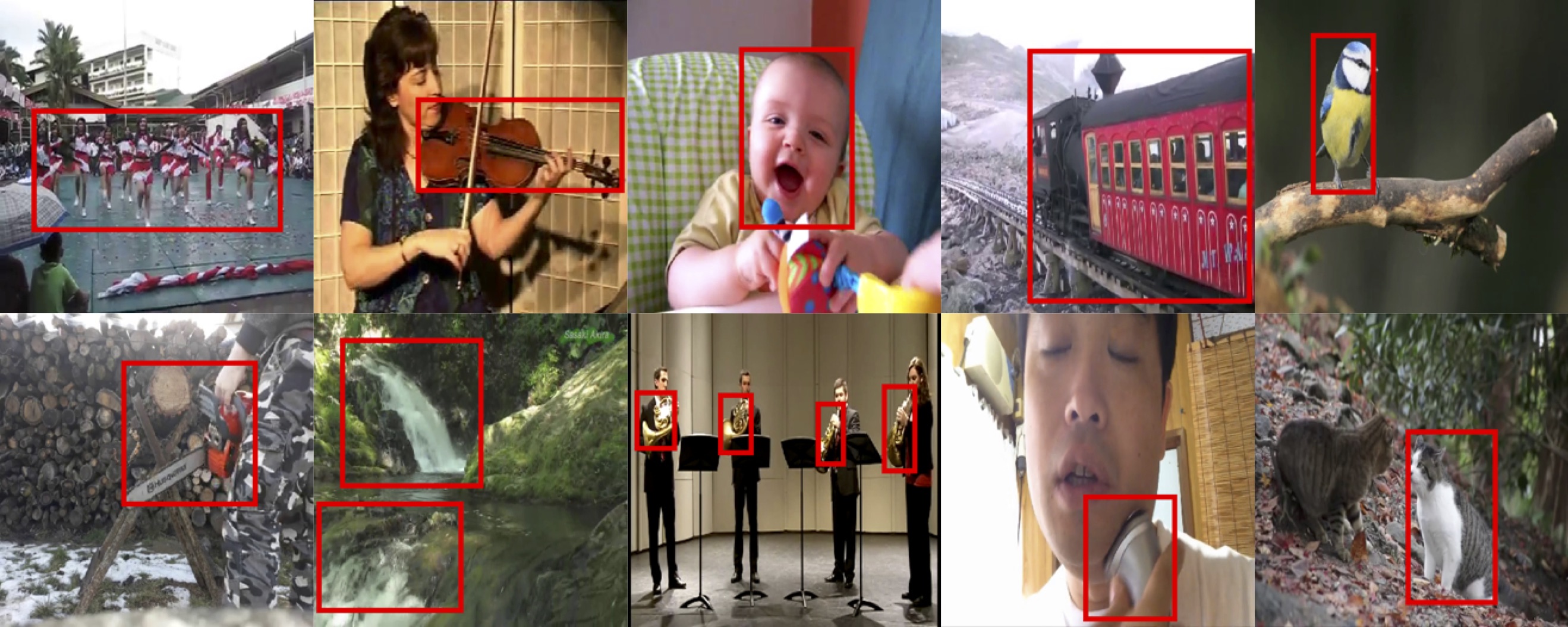}
  \caption{VGG-SS benchmark examples}
  \label{fig:sub1}
\end{subfigure}%
\begin{subfigure}{.25\textwidth}
  \centering
  \includegraphics[height=3cm]{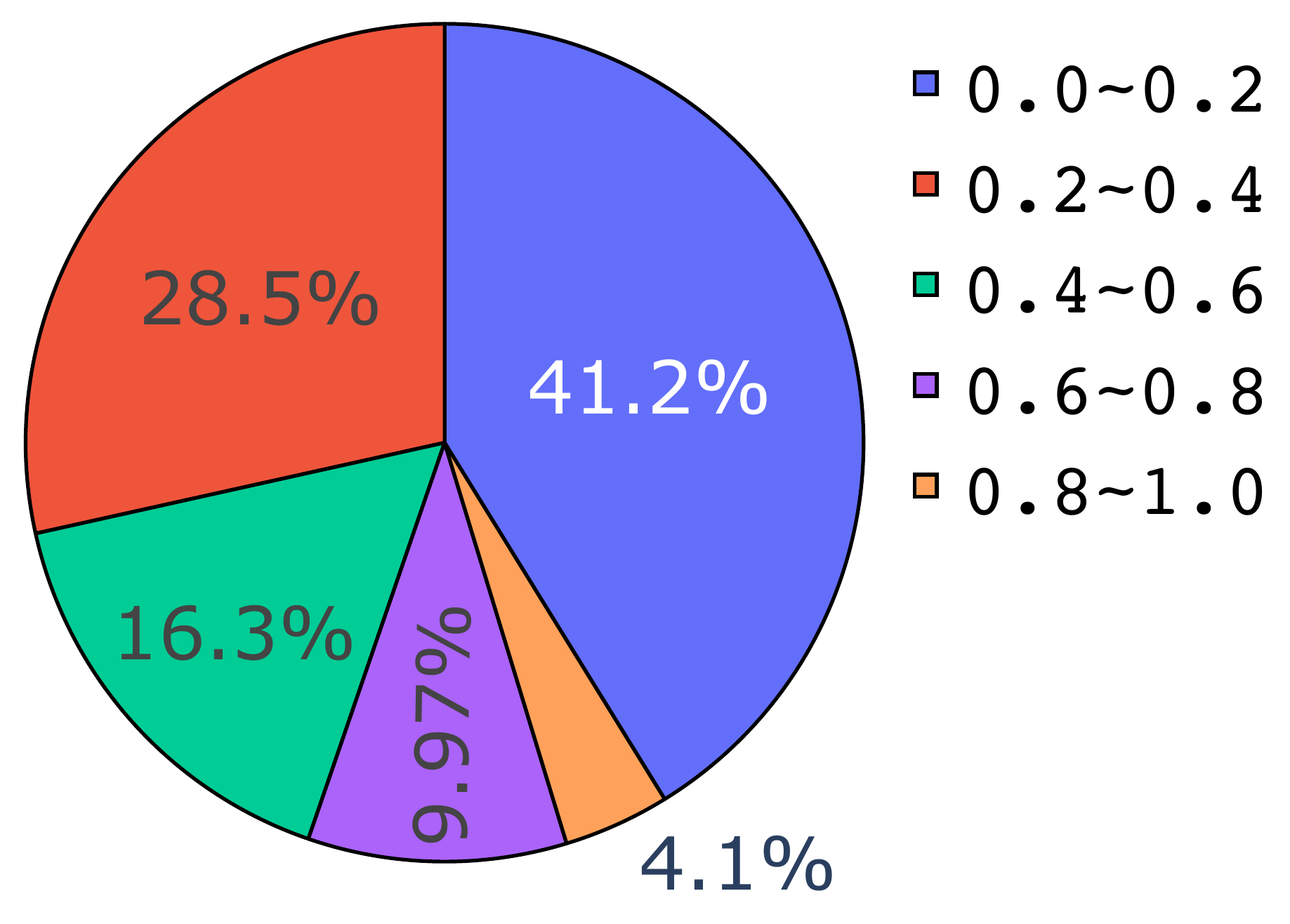}
  \caption{Bounding box areas}
  \label{fig:sub2}
\end{subfigure}
\begin{subfigure}{.25\textwidth}
  \centering
  \includegraphics[height=3cm]{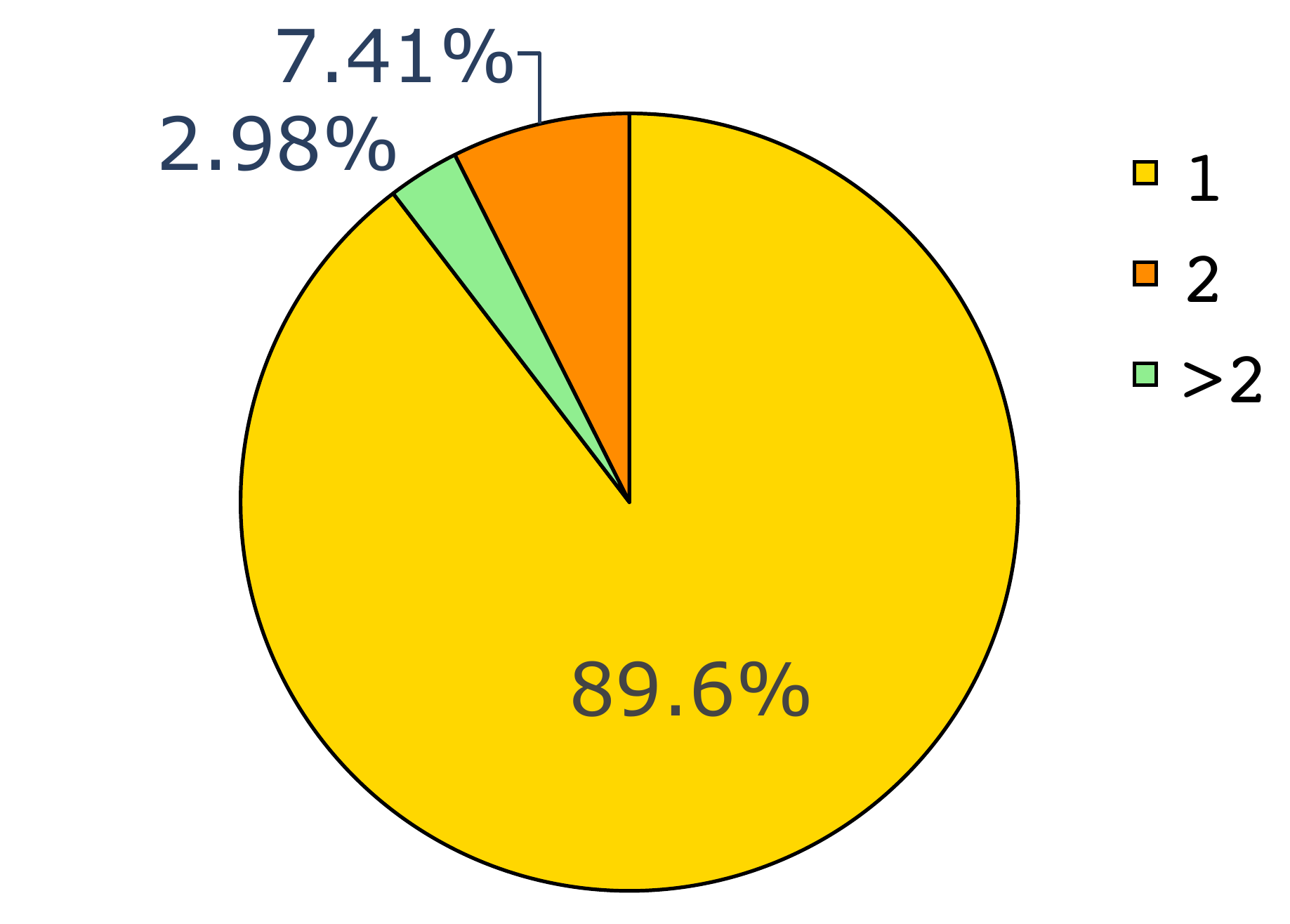}
  \caption{Number of bounding boxes}
  \label{fig:sub3}
\end{subfigure}
\caption{\textbf{VGG-SS Statistics.} ~\Cref{fig:sub1}: Example VGG-SS images and annotations showing class diversity (humans, animals, vehicles, tools etc.) ~\Cref{fig:sub2}: Distribution of bounding box areas in VGG-SS, the majority of boxes cover less than 40\% of the image area.~\Cref{fig:sub3} shows the distribution of number of bounding boxes - roughly 10\% of the test data is challenging with more than one bounding box per image.}
\label{fig:testset}
\end{figure*}



%% file: fig_table/Tab-testset_stat.tex
\begin{table}[!htb]
\vspace{1mm}
\begin{center}
\small
 \begin{tabular}{cccc} 
 \toprule
 Stage
 & Goal & \# Classes & \# Videos \\ \midrule 
 1 & Automatic BBox Generation & $309$ &  $15$k \\ 

 2 & Manual Annotation & $260$ & $8$k \\

 3 & Video Verification & $220$ & $5$k \\
 \bottomrule
\end{tabular}
\caption{The number of classes and videos in \dataset~after each annotation stage.}
\label{tab:stat_after_steps}
\vspace{-10mm}
\end{center}
\end{table}

%% file: section/experiments.tex
\section{Experiments}\label{sec:experiment}

In the following sections, we describe the datasets, evaluation protocol and experimental details used to thoroughly assess our method.

\subsection{Training Data}

For training our models, we consider two large-scale audio-visual datasets, the widely used Flickr SoundNet dataset and the recent VGG-Sound dataset, as detailed next.
Only the center frames of the \emph{raw} videos are used for training. Note, other frames~\eg (3/4 of the video) are tried for training, no considerable performance change is observed.\\[5pt]
\noindent\textbf{Flickr SoundNet:}
This dataset was initially proposed in~\cite{aytar16soundnet} and contains over 2 million unconstrained videos from Flickr. 
For a fair comparison with recent work~\cite{senocak2018learning,Hu_2019_CVPR,qian2020multiple}, we follow the same data splits, conducting self-supervised training with subsets of 10k or 144k image and audio pairs. \\[5pt]
%
\noindent\textbf{VGG-Sound:}
VGG-Sound was recently released with over 200k clips for 309 different sound categories. 
The dataset is conveniently audio-visual, in the sense that the object that emits sound is often visible in the corresponding video clip, which naturally suits the task considered in this paper. Again, to draw fair comparisons, 
we conduct experiments with training sets consisting of image and audio pairs of varying sizes, \ie~10k, 144k and the full set.

\subsection{Evaluation protocol}

In order to quantitatively evaluate the proposed approach, 
we adopt the evaluation metrics used in~\cite{senocak2018learning,qian2020multiple}:
Consensus Intersection over Union (cIoU) and Area Under Curve (AUC) are reported for each model on two test sets, as detailed next.\\[5pt]
\noindent\textbf{Flickr SoundNet Testset:} 
Following~\cite{senocak2018learning,Hu_2019_CVPR,qian2020multiple},
we report performance on the 250 annotated image-audio pairs of the Flickr SoundNet benchmark.
Every frame in this test set is accompanied by 20 seconds of audio, centered around it, and is annotated with 3 separate bounding boxes indicating the location of the sound source, each performed by a different annotator. \\[5pt]
\noindent\textbf{VGG-Sound Source~(VGG-SS):}
We also re-implement and train several baselines on VGG-Sound and evaluate them on our proposed VGG-SS benchmark, described in \cref{sec:dataset}. \\[-10pt]
\subsection{Implementation details}
As Flickr SoundNet consists of image-audio pairs, while VGG-Sound contains short video clips, when training on the latter we select the middle frame of the video clip and extract a 3s audio segment around it to create an equivalent image-audio pair.
Audio inputs are $257 \times 300$ magnitude spectrograms.
The dimensions for the audio output from the audio encoder CNN is a 512D vector,
which is max-pooled from a feature map of $17 \times 13 \times 512$,
where 17 and 13 refer to the frequency and time dimension respectively.
For the visual input, we resize the image to a $224  \times 224  \times 3$ tensor without cropping.
For both the visual and audio stream, we use a lightweight ResNet18~\cite{He16} as a backbone.
Following the baselines~\cite{Hu_2019_CVPR,qian2020multiple}, we also pretrain the visual encoder on ImageNet.
We use $\epsilon_p =0.65$ and $\epsilon_n=0.4$, $\tau = 0.03$,
that are picked by ablation study.
All models are trained with the Adam optimizer using a learning rate of $10^{-4}$ and a batch size of 256. 
During testing, we directly feed the full length audio spectrogram into the network. 

%% file: section/results.tex
\section{Results}

In the following sections, 
we first  compare our results with recent work on both Flickr SoundNet and VGG-SS dataset in detail.
Then we conduct an ablation analysis showing the importance of the \emph{hard negatives} and the Tri-map in self-supervised audio-visual localization.

\subsection{Comparison on the Flickr SoundNet Test Set}

In this section, we compare to recent approaches by training on the same amount of data (using various different datasets).
As shown in \Cref{Tab-MethodComp}, we first fix the training set to be Flickr SoundNet with 10k training samples and compare our method with~\cite{Arandjelovic18objects,qian2020multiple,harwath2018jointly}.
Our approach clearly outperforms the best previous methods by a substantial gap~(0.546\% vs.~0.582\%).
Second, we also train on VGG-Sound using 10k random samples, which shows the benefit of using VGG-Sound for training.
Third, we switch to a larger training set consisting of 144k samples, which gives us a further 5\% improvement compared to the previous state-of-the-art method~\cite{Hu_2019_CVPR}.
In order to tease apart the effect of various factors in our proposed approach, \ie~introducing \emph{hard negative} and using a Tri-map vs different training sets, \ie~Flickr144k vs.~VGG-Sound144k, we conduct an ablation study, as described next.

\input{fig_table/Tab-MethodComp}

\input{fig_table/Tab-Ablation}

\vspace{-10pt}
\subsection{Ablation Analysis}
In this section, 
we train our method using the 144k-samples training data from VGG-Sound and evaluate it on the Flickr SoundNet test set, 
as shown in \Cref{Tab.ablation}.\\
\par{\noindent \bf{On introducing hard negative and Tri-map.}}
While comparing \textbf{model~a} trained using only positives and \textbf{model~b} adding negatives from the complementary region decreases performance slightly.
This is because all the non-positive areas have been counted as negatives, 
whereas regions around the object are often hard to define.
Therefore deciding for all pixels whether they are positive or negative is problematic.
Second, comparing \textbf{model~b} and \textbf{model~c-f} where some areas between positives and negatives are ignored during training by using the Tri-map, 
we obtain a large gain~(around 2-4\%), 
demonstrating the importance of defining an ``uncertain'' region and allowing the model to self-tune. \\
\par{\noindent \bf{On hyperparameters.}}
we observe the model is generally robust to different set of hyper-parameters on defining the positive and negative regions,
\textbf{model-e}~($\epsilon_p =0.65$ and $\epsilon_n=0.4$) strives the best balance.

\input{fig_table/Tab-MethodComp2}
\input{fig_table/fig_examples}
\input{fig_table/Fig-visualisation}

\subsection{Comparison on VGG-Sound Source}

In this section,
we evaluate the models on the newly proposed VGG-SS benchmark.
As shown in \Cref{Tab-MethodComp2}, 
the CIoU is reduced significantly for all models compared to the results in \Cref{Tab-MethodComp}, showing that VGG-SS is a more diverse and challenging benchmark than Flickr SoundNet.
However, our proposed method still outperforms all other baseline methods by a large margin of around $5\%$.

\subsection{Qualitative results}

In \Cref{fig:trimap}, 
we threshold the heatmaps with different thresholds, \eg~$\epsilon_p =0.65$ and $\epsilon_n=0.4$~(same as the ones used during training). The objects and background are accurately highlighted in the positive region and negative region respectively, so that the model can learn proper amount of hard negatives.
We visualize the prediction results in \Cref{fig:vis},
and note that the proposed method presents much cleaner heatmap outputs.
This once again indicates the benefits of considering hard negatives during training.

\subsection{Open Set Audio-visual Localization}

We have so far trained and tested our models on data containing the same sound categories (closed set classification). 
In this section we determine if our model trained on heard/seen categories can generalize to classes that have never been heard/seen before, \ie to an open set scenario. 
To test this, we randomly sample 110 categories~(seen/heard) from VGG-Sound for training, 
and evaluate our network on another \textit{disjoint} set of 110 unseen/unheard categories 
(for a full list please refer to appendix).
We use roughly 70k samples for both heard and unheard classes.

\input{fig_table/Tab-Openset}

Heard and unheard evaluations are shown in \Cref{Tab-Openset}, where for the heard split we also train the model on 70k samples containing both old and new classes. The difference in performance is only 2\%, which demonstrates the ability of our network to generalize to unheard or unseen categories. This is not surprising due to the similarity between several categories. For example, if the training corpus contains human speech, one would expect the model to be capable of localizing human singing, as both classes share semantic similarities in audio and visual features.


%% file: fig_table/Tab-MethodComp.tex
\begin{table}[!htb]
\centering
\begin{tabular}{llcc}
\toprule
Method        						     	 & Training set  & CIoU    & AUC          \\\midrule
Attention10k~\cite{senocak2018learning}      & Flickr10k          & 0.436   & 0.449     \\
CoarsetoFine~\cite{qian2020multiple}         & Flickr10k          & 0.522   & 0.496 \\
AVObject~\cite{Afouras20b}                   & Flickr10k           & 0.546        & 0.504  \\
\textbf{Ours}         						 & Flickr10k      &   \textbf{0.582}      &  \textbf{0.525}\\ 
\textbf{Ours}         					     & VGG-Sound10k       & \textbf{0.618} & \textbf{0.536}  \\\hdashline
\\
Attention10k~\cite{senocak2018learning}      & Flickr144k      & 0.660  & 0.558 \\
DMC~\cite{Hu_2019_CVPR}      		         & Flickr144k       & 0.671   & 0.568  \\
\textbf{Ours}         					     & Flickr144k    & \textbf{0.699}   & \textbf{0.573}\\
\textbf{Ours}         					     & VGG-Sound144k    & \textbf{0.719}   & \textbf{0.582}\\
\textbf{Ours}         					     & VGG-Sound Full    & \textbf{0.735}  & \textbf{0.590}\\
\bottomrule

\end{tabular}%
{\vspace{-3pt}
\caption{Quantitative results on Flickr SoundNet testset. We outperform all recent works using different training sets and number of training data.}\label{Tab-MethodComp}}

\end{table}

%% file: fig_table/Tab-Ablation.tex
\begin{table}[!htb]
\centering
\vspace{-1em}
\begin{tabular}{clcclc}
\toprule
Model & Pos $\epsilon$  & Neg $\epsilon$ & Tri-map       & CIoU    & AUC \\ \midrule
a     & $\checkmark~(0.6)$  &  $\times$  & $\times$  & 0.675  & 0.568  \\
b     & $\checkmark~(0.6)$  & $\checkmark~(0.6)$  & $\times$  & 0.667  & 0.544 \\ \midrule
c     & $\checkmark~(0.6)$  &  $\checkmark~(0.45)$ & $\checkmark$  & 0.700  & 0.568 \\
d    & $\checkmark~(0.65)$  &  $\checkmark~(0.45)$ & $\checkmark$  & 0.703  & 0.569 \\
e    & $\checkmark~(0.65)$  &  $\checkmark~(0.4)$ & $\checkmark$  & \bf{0.719}  & \bf{0.582} \\
f     & $\checkmark~(0.7)$  &  $\checkmark~(0.3)$ & $\checkmark$  & 0.687  & 0.563 \\
\bottomrule
\end{tabular}%
{\caption{Ablation study.
We investigate the effects of the hyper-parameters for defining positive and negative regions,
where the picked value is specified in the bracket.}\label{Tab.ablation}}
\end{table}

%% file: fig_table/Tab-MethodComp2.tex
\begin{table}[t]
\centering
\begin{tabular}{lcc}
\toprule
Method        						    & CIoU    & AUC \\ \midrule
Attention10k~\cite{senocak2018learning}    &  0.185 & 0.302  \\
AVobject~\cite{Afouras20b}     		       & 0.297 & 0.357 \\
\textbf{Ours}         					     & \textbf{0.344} & \textbf{0.382} \\
\bottomrule
\end{tabular}%
{\vspace{-3pt}
\caption{Quantitative results on the VGG-SS testset. All models are trained on VGG-Sound 144k and tested on VGG-SS.}\label{Tab-MethodComp2}}
\end{table}

%% file: fig_table/fig_examples.tex
\begin{figure*}[!htb]
\centering
\includegraphics[width=0.98\textwidth]{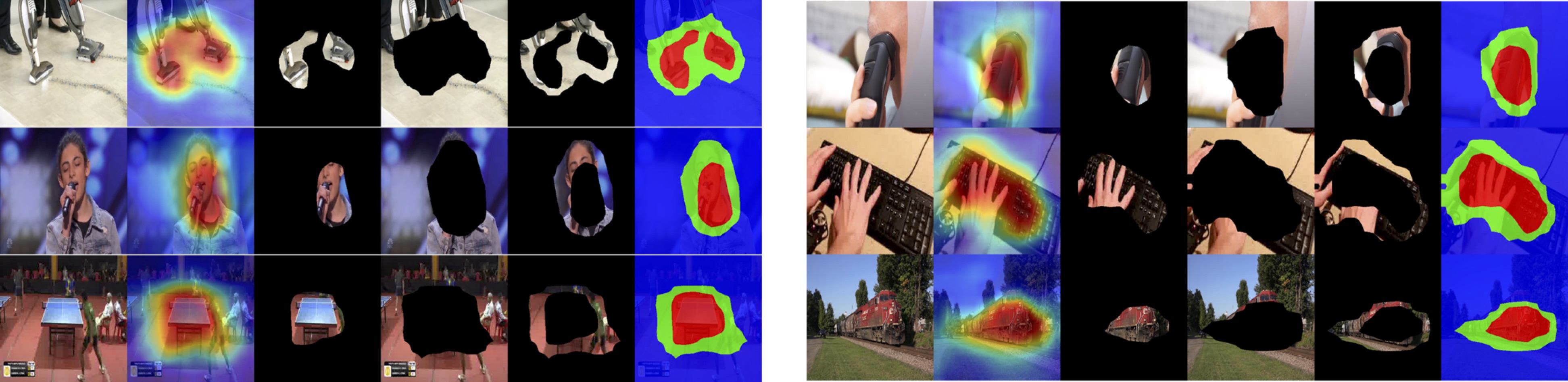}
\vspace{-5pt}
\caption{\textbf{Example Tri-map visualisations.} We show images, heatmaps and Tri-maps here. The Tri-map effectively identify the objects and the uncertain region let the model only learn controlled hard negatives. 
}
\label{fig:trimap}
\end{figure*}

%% file: fig_table/Fig-visualisation.tex
\begin{figure*}[!htb]
\centering
\begin{subfigure}{.49\textwidth}
  \centering
  \includegraphics[width=.98\linewidth]{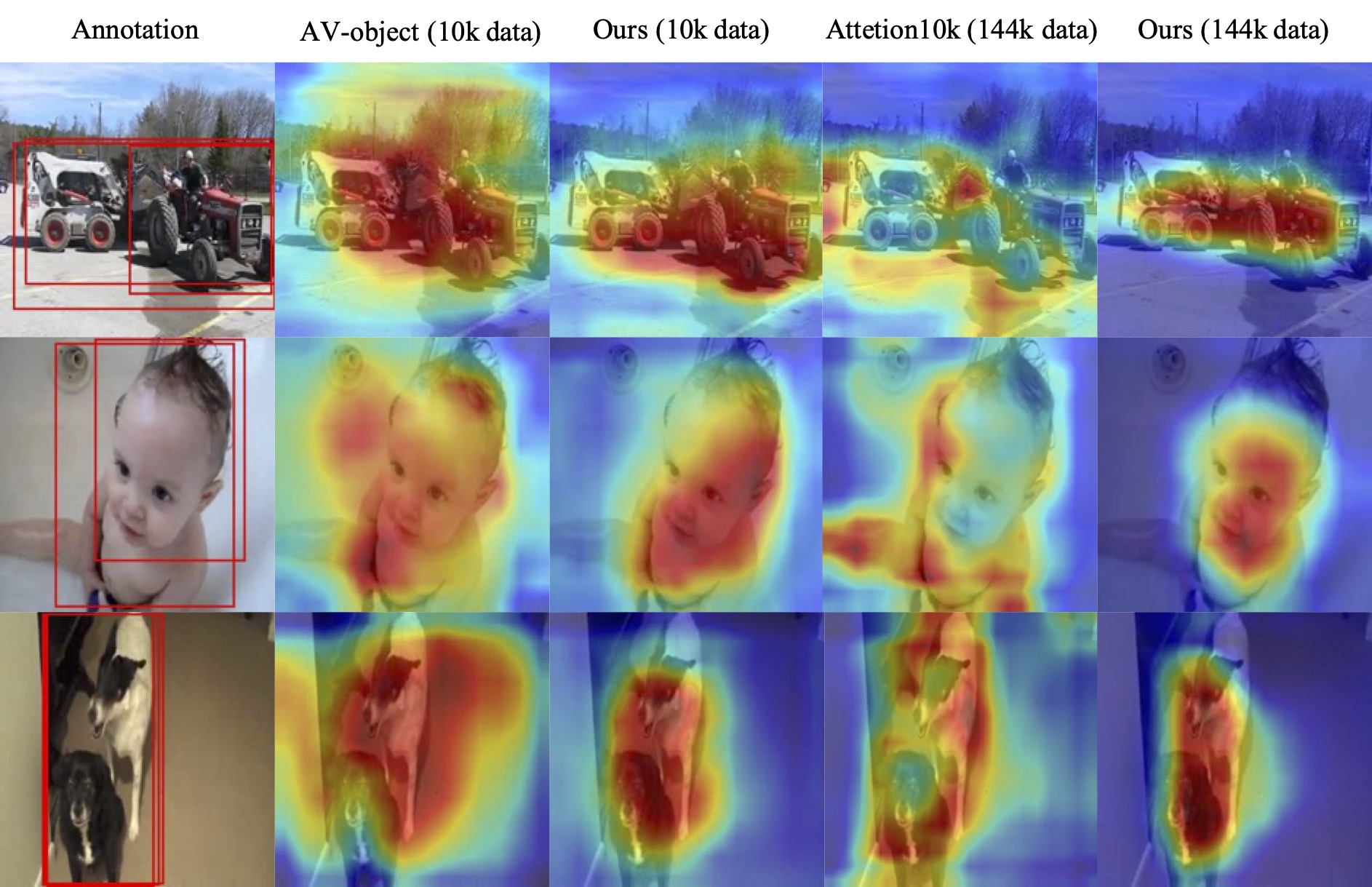}
  \caption{Visualisation on Flickr SoundNet testset}
  \label{fig:flickr_vis}
\end{subfigure}%
\begin{subfigure}{.49\textwidth}
  \centering
  \includegraphics[width=.98\linewidth]{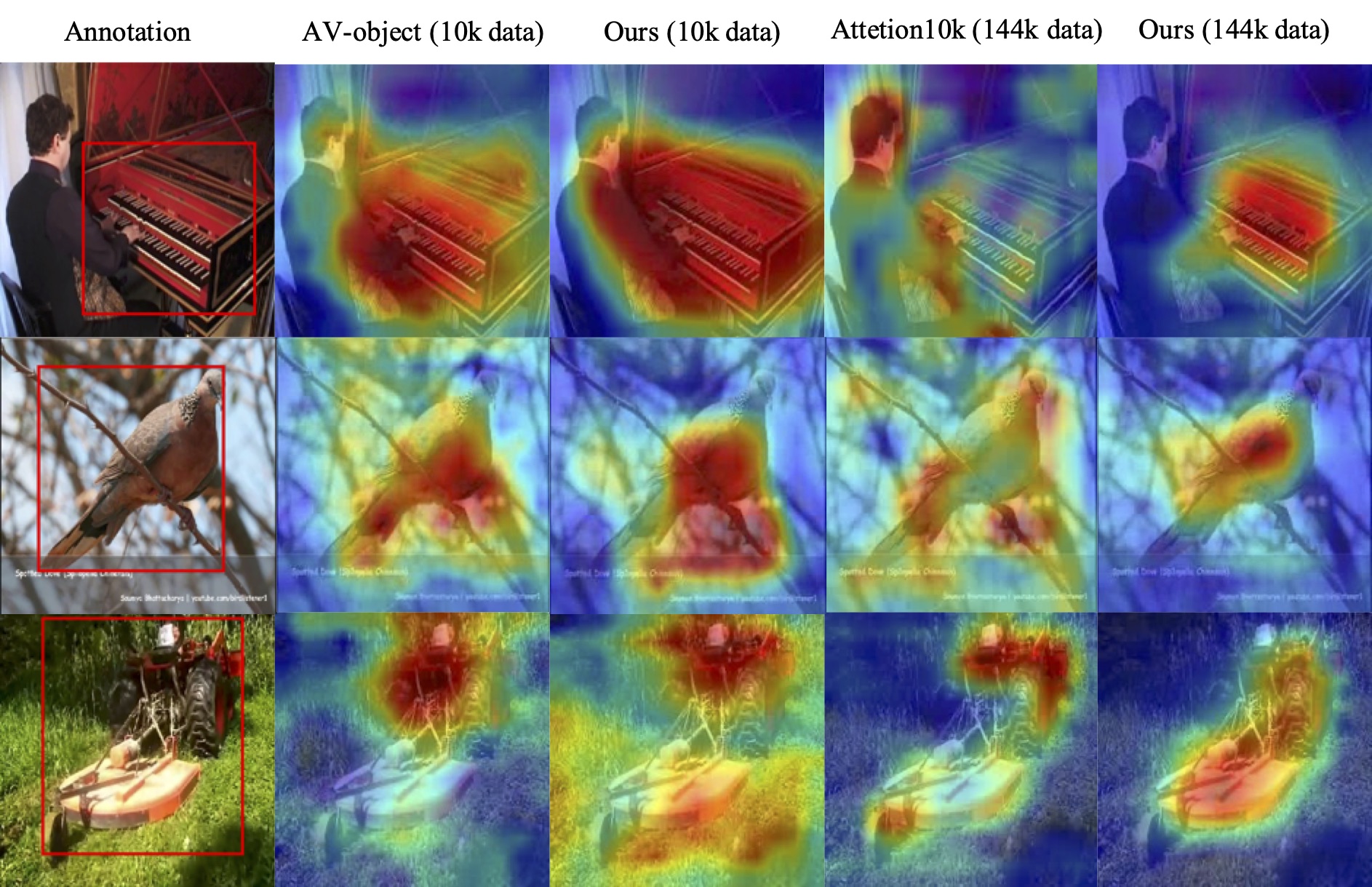}
  \caption{Visualisation on VGG-SS testset}
  \label{fig:vgg_vis}
\end{subfigure}
\vspace{-3pt}
\caption{\textbf{Qualitative results} for models trained on various methods and data amount. The first column shows annotation overlaid on images, the following two column shows predictions trained on 10k data and the last tow column show predictions trained on 144k data. Our method has no false positives in the predictions as the hard negatives are penalised in the training. }
\label{fig:vis}
\end{figure*}

%% file: fig_table/Tab-Openset.tex
\begin{table}[t]
\centering
\vspace{1em}
\begin{tabular}{lrcc}
\toprule
\# training Data       & Test class 						    & CIoU    & AUC \\ \midrule
70k    		 & Heard 110      & 0.289 & 0.362 \\
70k          & Unheard 110  	& 0.263 & 0.347 \\
\bottomrule
\end{tabular}%
{\vspace{-3pt}
\caption{Quantitative results on VGG-SS for unheard classes. We vary the training set (classes) and keep the testing set fixed (subset of the VGG-SS).}\label{Tab-Openset}}
\vspace{-10pt}
\end{table}

%% file: section/conclusion.tex
\section{Conclusion}\label{sec:conclusion}

We revisit the problem of unsupervised visual sound source localization.
For this task, we introduce a new large-scale benchmark called VGG-Sound Source, which is more challenging than existing ones such as Flickr SoundNet.
We also suggest a simple, general and effective technique that significantly boosts the performance of existing sound source locators, by explicitly mining for hard negative image locations in the same image that contains the sounding objecs.
A careful implementation of this idea using Tri-maps and differentiable thresholding allows us to significantly outperform the state of the art.

%% file: section/appendix.tex
\begin{appendices}
\section{Evaluation metric}
We follow the same evaluation metrics as in~\cite{senocak2018learning}, and report consensus intersection over union (cIoU) and area under curve (AUC). The Flickr Soundnet dataset contains 3 bounding box annotations from different human annotators. The bounding box annotations are first converted into binary masks $\{\mathbf{b}_j\}_{j=1}^n$ 
where $n$ is the number of bounding box annotations per image. The final weighted ground truth mask is defined as:
\begin{align*}
\centering
\mathbf{g} = min(\sum_{j=1}^{n}\frac{\mathbf{b}_j}{C},1)
\end{align*}
where $C$ is a parameter meaning the minimum number of opinions to reach agreement. We choose $C=2$, the same as~\cite{senocak2018learning}. Given the ground truth $\mathbf{g}$ and our prediction $\mathbf{p}$,
the cIoU is defined as 
\begin{align*}
\centering
cIoU(\tau) = \frac{\sum_{i \in A(\tau)}g_i}{\sum_i{g_i} + \sum_{i \in A(\tau) -G}1}
\end{align*}
where i indicates the pixel index of the map, $\tau$ denotes the threshold to judge positiveness, $A(\tau) = \{ i|p_i > \tau\}$, and $G = \{ i|g_i>0\}$. We follow~\cite{senocak2018learning}, and use $\tau = 0.5$. Example predictions and their cIoUs are shown in~\Cref{fig:ciou}.

\input{fig_table/ciou}

Since the $cIoU$ is calculated for each testing image-audio pair, the success ratio is defined as number of successful samples ($cIoU$ greater than a threshold $\tau_2$) / total number of samples. The curve showing success ratio is plotted against the threshold $\tau_2$ varied from 0 to 1 and the area under the curve is reported. The Pseudocode is shown in~\Cref{alg:code}.

\begin{algorithm}[!htb]
\caption{Pseudocode of AUC calculation}
\label{alg:code}
\algcomment{\fontsize{7.2pt}{0em}\selectfont \texttt{bmm}: batch matrix multiplication; \texttt{mm}: matrix multiplication; \texttt{cat}: concatenation.
}
\definecolor{codeblue}{rgb}{0.25,0.5,0.5}
\lstset{
  backgroundcolor=\color{white},
  basicstyle=\fontsize{7.2pt}{7.2pt}\ttfamily\selectfont,
  columns=fullflexible,
  breaklines=true,
  captionpos=b,
  commentstyle=\fontsize{7.2pt}{7.2pt}\color{codeblue},
  keywordstyle=\fontsize{7.2pt}{7.2pt},
}
\begin{lstlisting}[language=python]
# cIoUs : [cIoU_1,cIoU_2,...,cIoU_n]

x = [0.05 * i for i in range(21)]
for t in x: # Divide into 20 different thresholds
	score.append(sum(cIoUs > t) / len(cIoUs))
AUC = calulcate_auc(x,score) # sklearn.metrics.auc
\end{lstlisting}
\end{algorithm}

\subsection{Tri-map visualisation} 
In addition to video examples, we show more image results of our Tri-maps in~\Cref{fig:trimap2}.
\input{fig_table/trimap}

\clearpage

\section{VGG-Sound Source (VGG-SS)}
We show more dataset examples, the full 220 class list of VGG-SS and the classes we removed from the original VGG-Sound dataset~\cite{Chen20a} in this section.


\subsection{VGG-SS annotation interface} 
We show our manual annotation interface, LISA~\cite{Dutta19a}, in~\Cref{fig:interface}. 
The example videos are from the class `Rapping'. 
The `Play' button shows the 5s clip, and `Show region' recenters to the key frame we want to annotate. 
We choose `Yes' only if we hear the correct sound, `No' for the clips that do not contain the sound of class, 
and `Not Sure' if  the sound is not within the 5s we choose~(original video clip is 10s) 
\input{fig_table/interface}

\subsection{VGG-SS examples}
\input{fig_table/vggss_example1}

We randomly sample from images with 1 bounding box, 2 bounding boxes, and with more than 2 bounding boxes. We show examples with 1 bounding box on the top 4 rows, examples with 2 bounding boxes on the following two rows, and examples with more than 2 bounding boxes on the last row in~\Cref{fig:vggss}.
\clearpage
\input{fig_table/vgg_bar}
\twocolumn
\subsection{VGG-SS class list} 
We show a bar chart of per-class frequencies for the VGG-SS testset in~\Cref{fig:bar}.
The full list below is shown in the format of \emph{index. class name (number of clips in the class)}.
\input{fig_table/vggss_classlist}

\subsection{Removed classes} 
\input{fig_table/vggss_classlist_remove}
\end{appendices}

%% file: fig_table/ciou.tex
\begin{figure}[!htb]
\centering
\includegraphics[width=0.8\textwidth]{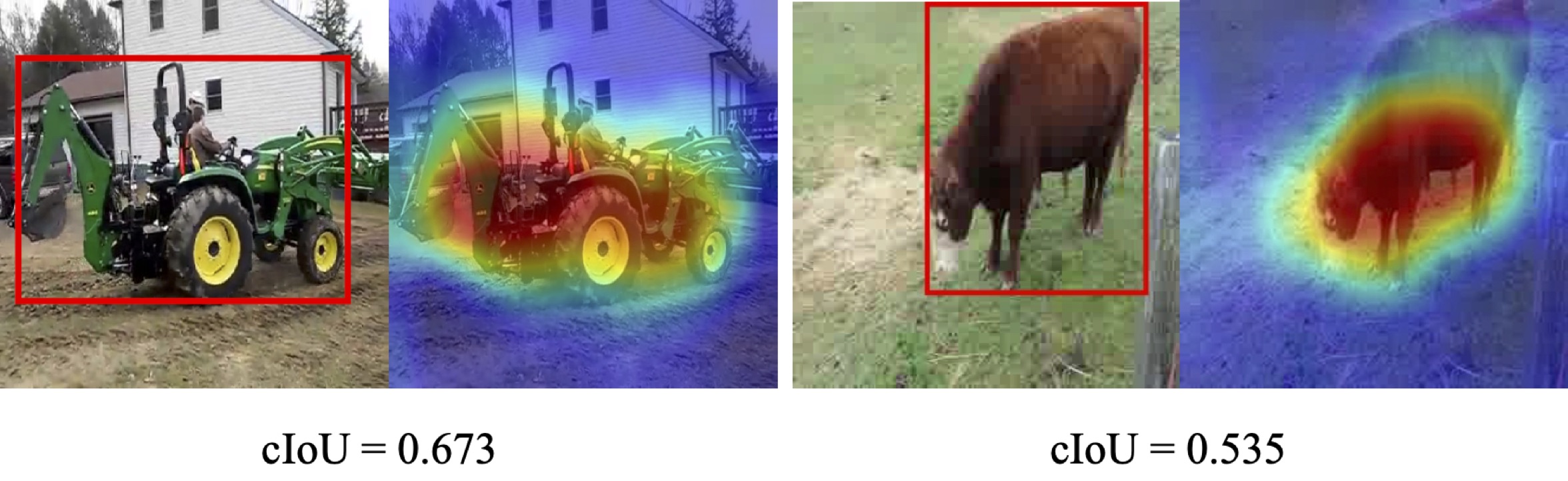}
\caption{Example predictions with calculated cIoU.}\label{fig:ciou}
\end{figure}

%% file: fig_table/trimap.tex
\begin{figure*}
\includegraphics[width=1\textwidth]{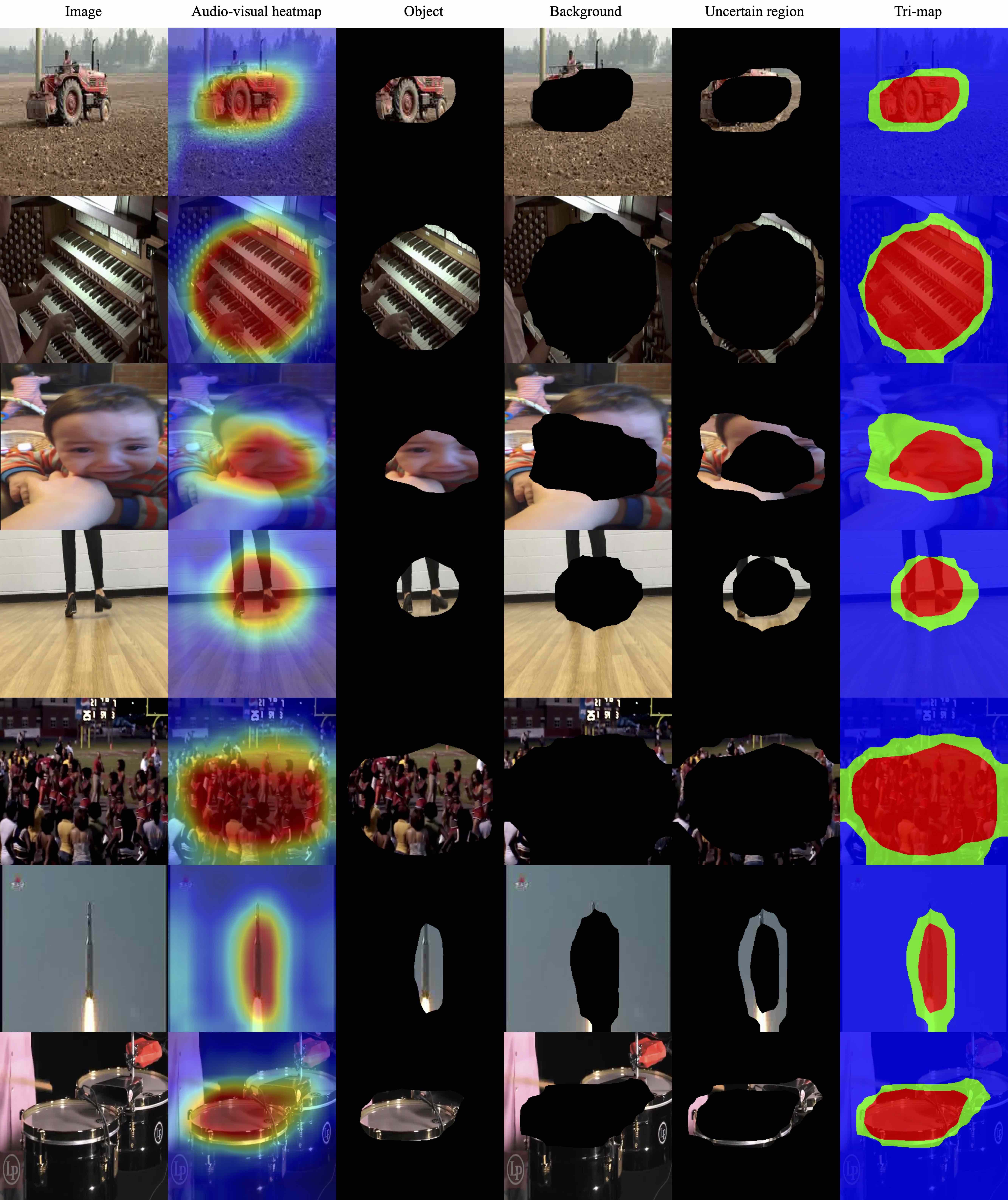}
\caption{Tri-map visualization examples.}\label{fig:trimap2}
\end{figure*}

%% file: fig_table/interface.tex
\begin{figure*}[!htb]
\includegraphics[width=1\textwidth]{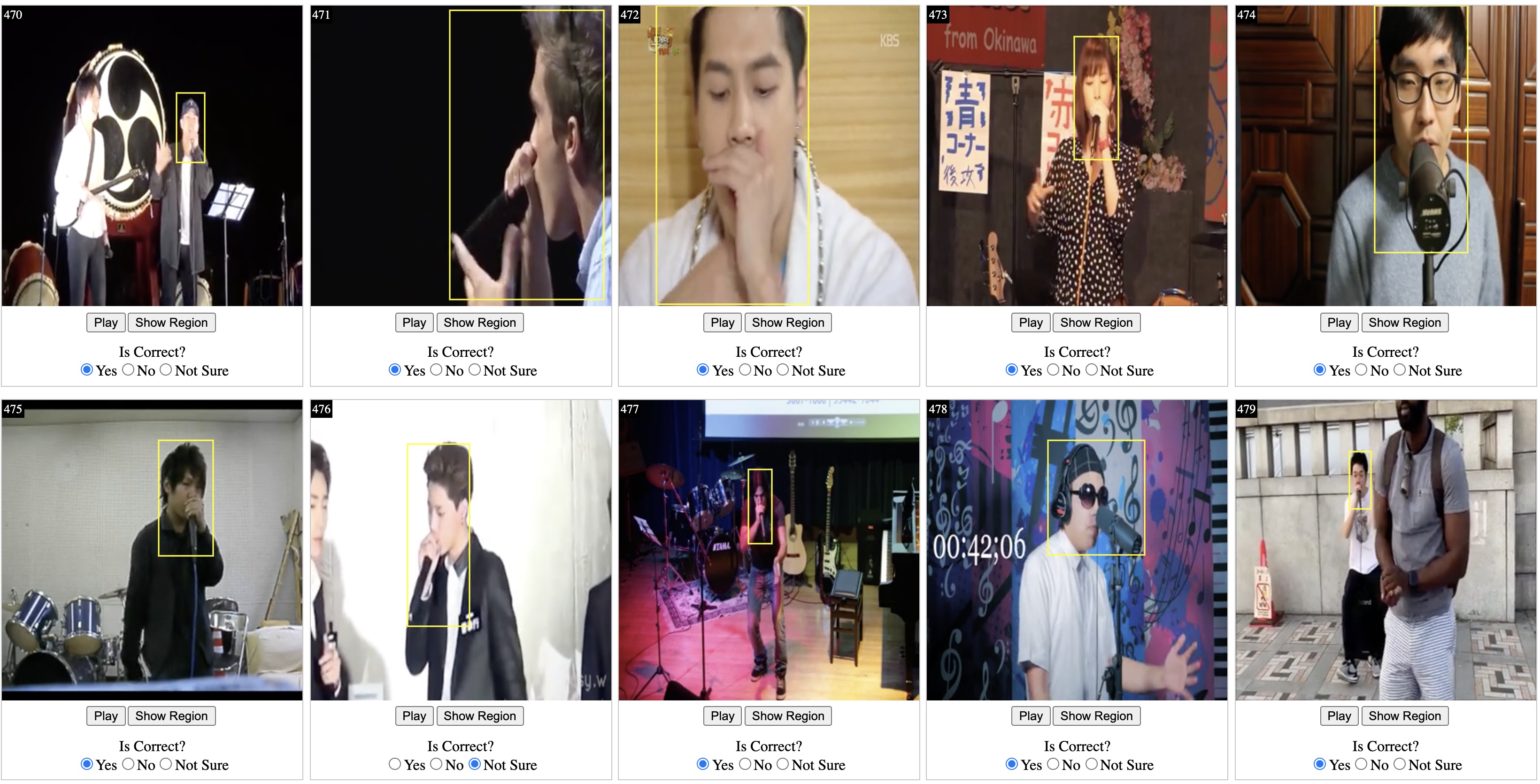}
\caption{LISA Annotation Interface}
\label{fig:interface}
\end{figure*}

%% file: fig_table/vggss_example1.tex
\begin{figure*}
\includegraphics[width=\textwidth]{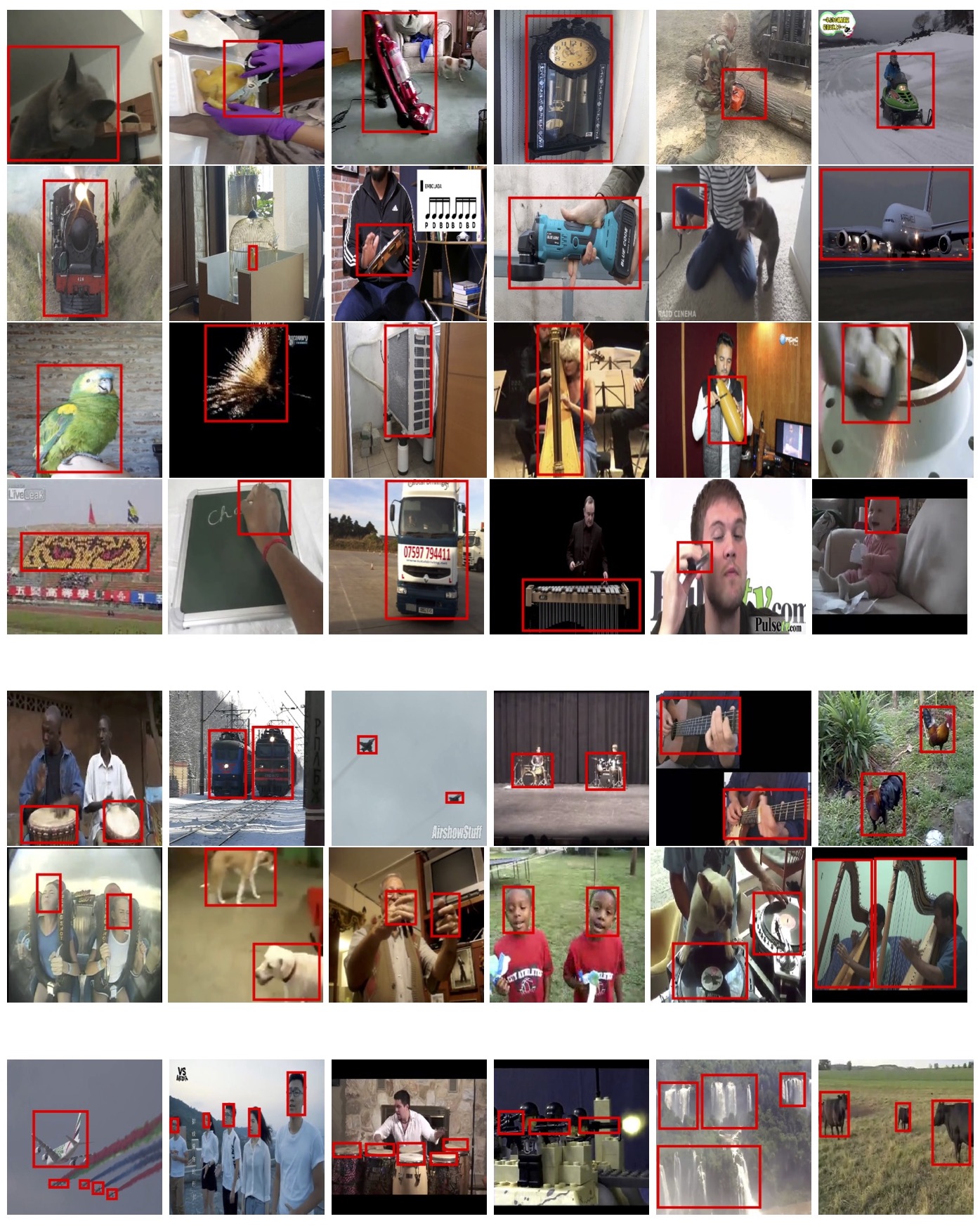}
\caption{We show examples with 1 bounding box on the top 4 rows, examples with 2 bounding boxes on the following two rows, and examples with more than 2 bounding boxes on the last row.}\label{fig:vggss}
\end{figure*}

%% file: fig_table/vgg_bar.tex
\begin{figure*}
\centering
\includegraphics[height=1\textheight]{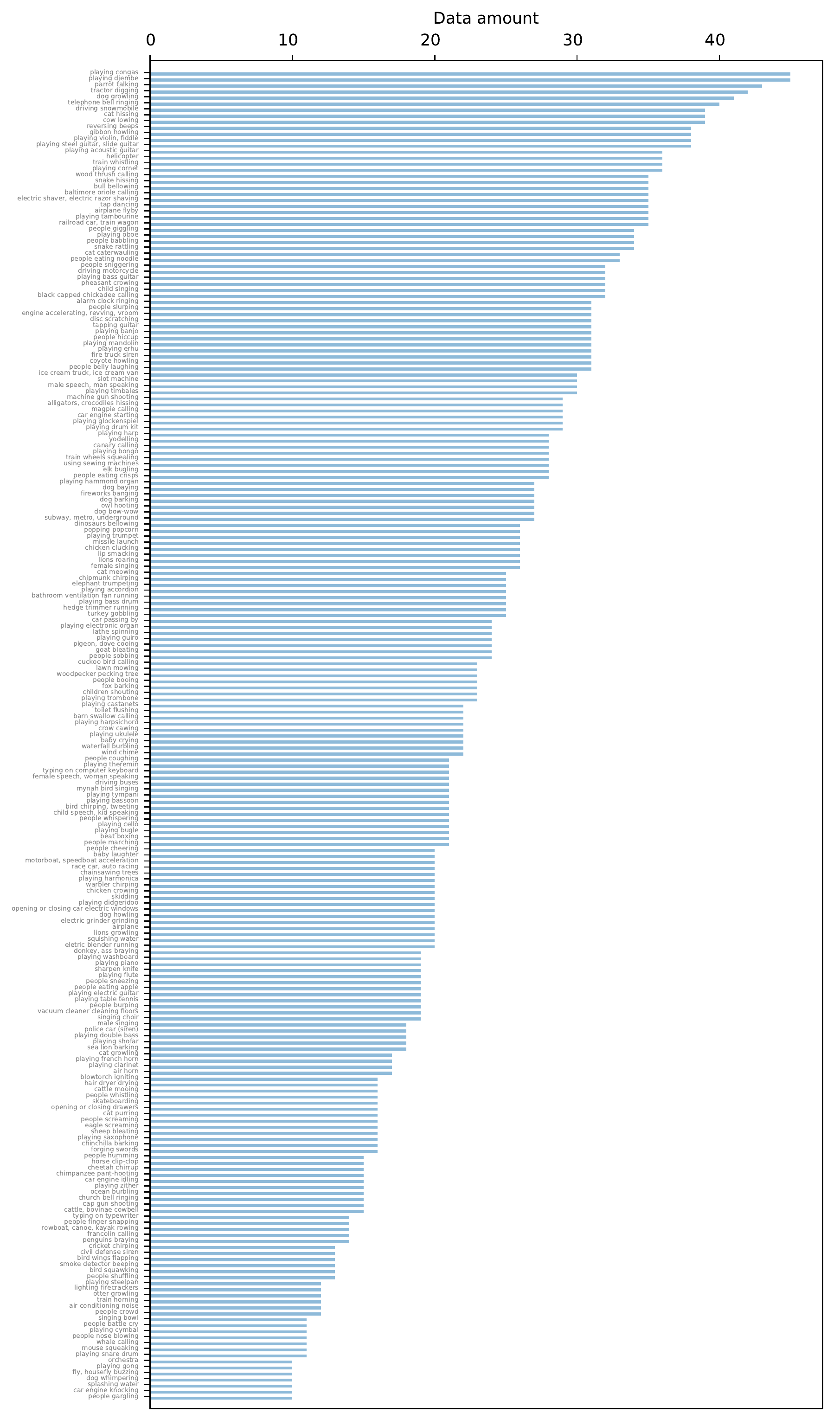}
\caption{VGG-SS benchmark per class statistics.}\label{fig:bar}
\end{figure*}

%% file: fig_table/vggss_classlist.tex
\begin{enumerate}
\item playing congas (45)
\item playing djembe (45)
\item parrot talking (43)
\item tractor digging (42)
\item dog growling (41)
\item telephone bell ringing (40)
\item driving snowmobile (39)
\item cat hissing (39)
\item cow lowing (39)
\item reversing beeps (38)
\item gibbon howling (38)
\item playing violin, fiddle (38)
\item playing steel guitar, slide guitar (38)
\item playing acoustic guitar (36)
\item helicopter (36)
\item train whistling (36)
\item playing cornet (36)
\item wood thrush calling (35)
\item snake hissing (35)
\item bull bellowing (35)
\item baltimore oriole calling (35)
\item electric shaver, electric razor shaving (35)
\item tap dancing (35)
\item airplane flyby (35)
\item playing tambourine (35)
\item railroad car, train wagon (35)
\item people giggling (34)
\item playing oboe (34)
\item people babbling (34)
\item snake rattling (34)
\item cat caterwauling (33)
\item people eating noodle (33)
\item people sniggering (32)
\item driving motorcycle (32)
\item playing bass guitar (32)
\item pheasant crowing (32)
\item child singing (32)
\item black capped chickadee calling (32)
\item alarm clock ringing (31)
\item people slurping (31)
\item engine accelerating, revving, vroom (31)
\item disc scratching (31)
\item tapping guitar (31)
\item playing banjo (31)
\item people hiccup (31)
\item playing mandolin (31)
\item playing erhu (31)
\item fire truck siren (31)
\item coyote howling (31)
\item people belly laughing (31)
\item ice cream truck, ice cream van (30)
\item slot machine (30)
\item male speech, man speaking (30)
\item playing timbales (30)
\item machine gun shooting (29)
\item alligators, crocodiles hissing (29)
\item magpie calling (29)
\item car engine starting (29)
\item playing glockenspiel (29)
\item playing drum kit (29)
\item playing harp (28)
\item yodelling (28)
\item canary calling (28)
\item playing bongo (28)
\item train wheels squealing (28)
\item using sewing machines (28)
\item elk bugling (28)
\item people eating crisps (28)
\item playing hammond organ (27)
\item dog baying (27)
\item fireworks banging (27)
\item dog barking (27)
\item owl hooting (27)
\item dog bow-wow (27)
\item subway, metro, underground (27)
\item dinosaurs bellowing (26)
\item popping popcorn (26)
\item playing trumpet (26)
\item missile launch (26)
\item chicken clucking (26)
\item lip smacking (26)
\item lions roaring (26)
\item female singing (26)
\item cat meowing (25)
\item chipmunk chirping (25)
\item elephant trumpeting (25)
\item playing accordion (25)
\item bathroom ventilation fan running (25)
\item playing bass drum (25)
\item hedge trimmer running (25)
\item turkey gobbling (25)
\item car passing by (24)
\item playing electronic organ (24)
\item lathe spinning (24)
\item playing guiro (24)
\item pigeon, dove cooing (24)
\item goat bleating (24)
\item people sobbing (24)
\item cuckoo bird calling (23)
\item lawn mowing (23)
\item woodpecker pecking tree (23)
\item people booing (23)
\item fox barking (23)
\item children shouting (23)
\item playing trombone (23)
\item playing castanets (22)
\item toilet flushing (22)
\item barn swallow calling (22)
\item playing harpsichord (22)
\item crow cawing (22)
\item playing ukulele (22)
\item baby crying (22)
\item waterfall burbling (22)
\item wind chime (22)
\item people coughing (21)
\item playing theremin (21)
\item typing on computer keyboard (21)
\item female speech, woman speaking (21)
\item driving buses (21)
\item mynah bird singing (21)
\item playing tympani (21)
\item playing bassoon (21)
\item bird chirping, tweeting (21)
\item child speech, kid speaking (21)
\item people whispering (21)
\item playing cello (21)
\item playing bugle (21)
\item beat boxing (21)
\item people marching (21)
\item people cheering (20)
\item baby laughter (20)
\item motorboat, speedboat acceleration (20)
\item race car, auto racing (20)
\item chainsawing trees (20)
\item playing harmonica (20)
\item warbler chirping (20)
\item chicken crowing (20)
\item skidding (20)
\item playing didgeridoo (20)
\item opening or closing car electric windows (20)
\item dog howling (20)
\item electric grinder grinding (20)
\item airplane (20)
\item lions growling (20)
\item squishing water (20)
\item eletric blender running (20)
\item donkey, ass braying (19)
\item playing washboard (19)
\item playing piano (19)
\item sharpen knife (19)
\item playing flute (19)
\item people sneezing (19)
\item people eating apple (19)
\item playing electric guitar (19)
\item playing table tennis (19)
\item people burping (19)
\item vacuum cleaner cleaning floors (19)
\item singing choir (19)
\item male singing (18)
\item police car (siren) (18)
\item playing double bass (18)
\item playing shofar (18)
\item sea lion barking (18)
\item cat growling (17)
\item playing french horn (17)
\item playing clarinet (17)
\item air horn (17)
\item blowtorch igniting (16)
\item hair dryer drying (16)
\item cattle mooing (16)
\item people whistling (16)
\item skateboarding (16)
\item opening or closing drawers (16)
\item cat purring (16)
\item people screaming (16)
\item eagle screaming (16)
\item sheep bleating (16)
\item playing saxophone (16)
\item chinchilla barking (16)
\item forging swords (16)
\item people humming (15)
\item horse clip-clop (15)
\item cheetah chirrup (15)
\item chimpanzee pant-hooting (15)
\item car engine idling (15)
\item playing zither (15)
\item ocean burbling (15)
\item church bell ringing (15)
\item cap gun shooting (15)
\item cattle, bovinae cowbell (15)
\item typing on typewriter (14)
\item people finger snapping (14)
\item rowboat, canoe, kayak rowing (14)
\item francolin calling (14)
\item penguins braying (14)
\item cricket chirping (13)
\item civil defense siren (13)
\item bird wings flapping (13)
\item smoke detector beeping (13)
\item bird squawking (13)
\item people shuffling (13)
\item playing steelpan (12)
\item lighting firecrackers (12)
\item otter growling (12)
\item train horning (12)
\item air conditioning noise (12)
\item people crowd (12)
\item singing bowl (11)
\item people battle cry (11)
\item playing cymbal (11)
\item people nose blowing (11)
\item whale calling (11)
\item mouse squeaking (11)
\item playing snare drum (11)
\item orchestra (10)
\item playing gong (10)
\item fly, housefly buzzing (10)
\item dog whimpering (10)
\item splashing water (10)
\item car engine knocking (10)
\item people gargling (10)
\end{enumerate}

%% file: fig_table/vggss_classlist_remove.tex
\begin{enumerate}
\item running electric fan
\item mouse clicking
\item people eating
\item people clapping
\item roller coaster running
\item cell phone buzzing
\item basketball bounce
\item playing timpani
\item people running
\item firing muskets
\item door slamming
\item hammering nails
\item chopping wood
\item striking bowling
\item bowling impact
\item ripping paper
\item baby babbling
\item playing hockey
\item swimming
\item hail
\item people slapping
\item wind rustling leaves
\item sea waves
\item heart sounds, heartbeat
\item raining
\item rope skipping
\item stream burbling
\item playing badminton
\item striking pool
\item wind noise
\item bouncing on trampoline
\item thunder
\item ice cracking
\item shot football
\item playing squash
\item scuba diving
\item cupboard opening or closing
\item fire crackling
\item playing volleyball
\item golf driving
\item sloshing water
\item sliding door
\item playing tennis
\item footsteps on snow
\item people farting
\item playing marimba, xylophone
\item foghorn
\item tornado roaring
\item playing lacrosse
\end{enumerate}

%% file: main.bbl
\begin{thebibliography}{10}\itemsep=-1pt

\bibitem{Afouras20b}
Triantafyllos Afouras, Andrew Owens, Joon~Son Chung, and Andrew Zisserman.
\newblock Self-supervised learning of audio-visual objects from video.
\newblock In {\em Proc. ECCV}, 2020.

\bibitem{Arandjelovic18objects}
Relja Arandjelovic and Andrew Zisserman.
\newblock Objects that sound.
\newblock In {\em Proc. ECCV}, 2017.

\bibitem{aytar16soundnet}
Yusuf Aytar, Carl Vondrick, and Antonio Torralba.
\newblock Soundnet: Learning sound representations from unlabeled video.
\newblock In {\em NeurIPS}, 2016.

\bibitem{Chen20a}
Honglie Chen, Weidi Xie, Andrea Vedaldi, and Andrew Zisserman.
\newblock V{GG-S}ound: A large-scale audio-visual dataset.
\newblock In {\em Proc. ICASSP}, 2020.

\bibitem{Chuang02}
Yung-Yu Chuang, Aseem Agarwala, Brian Curless, David~H. Salesin, and Richard
  Szeliski.
\newblock Video matting of complex scenes.
\newblock {\em ACM Trans. Graph}, 2002.

\bibitem{chung16}
Joon~Son Chung and Andrew Zisserman.
\newblock Lip reading in the wild.
\newblock In {\em Proc. ACCV}, 2016.

\bibitem{dalal05histogram}
Navneet Dalal and Bill Triggs.
\newblock Histograms of oriented gradients for human detection.
\newblock In {\em Proc. {CVPR}}, 2005.

\bibitem{Dutta19a}
Abhishek Dutta and Andrew Zisserman.
\newblock The via annotation software for images, audio and video.
\newblock In {\em Proc. ACMM}, 2019.

\bibitem{fisher2000learning}
John~W Fisher~III, Trevor Darrell, William~T Freeman, and Paul~A Viola.
\newblock Learning joint statistical models for audio-visual fusion and
  segregation.
\newblock In {\em NeurIPS}, 2000.

\bibitem{gan2019self}
Chuang Gan, Hang Zhao, Peihao Chen, David Cox, and Antonio Torralba.
\newblock Self-supervised moving vehicle tracking with stereo sound.
\newblock In {\em Proc. ICCV}, 2019.

\bibitem{gao2019visual}
Ruohan Gao and Kristen Grauman.
\newblock 2.5d visual sound.
\newblock In {\em Proc. CVPR}, 2019.

\bibitem{Gemmeke17}
J Gemmeke, D Ellis, D Freedman, A Jansen, W Lawrence, C Moore, M Plakal, and M
  Ritter.
\newblock Audio {S}et: An ontology and human-labeled dataset for audio events.
\newblock In {\em Proc. ICASSP}, 2017.

\bibitem{Girshick14}
Ross Girshick, Jeff Donahue, Trevor Darrell, and Jitendra Malik.
\newblock Rich feature hierarchies for accurate object detection and semantic
  segmentation.
\newblock In {\em Proc. CVPR}, 2014.

\bibitem{Han19}
Tengda Han, Weidi Xie, and Andrew Zisserman.
\newblock Video representation learning by dense predictive coding.
\newblock In {\em Workshop on Large Scale Holistic Video Understanding, ICCV},
  2019.

\bibitem{harwath2018jointly}
David Harwath, Adria Recasens, D{\'\i}dac Sur{\'\i}s, Galen Chuang, Antonio
  Torralba, and James Glass.
\newblock Jointly discovering visual objects and spoken words from raw sensory
  input.
\newblock In {\em Proc. ECCV}, 2018.

\bibitem{He16}
Kaiming He, Xiangyu Zhang, Shaoqing Ren, and Jian Sun.
\newblock Deep residual learning for image recognition.
\newblock In {\em Proc. CVPR}, 2016.

\bibitem{hershey1999audio}
John~R. Hershey and Javier~R. Movellan.
\newblock Audio-vision: Locating sounds via audio-visual synchrony.
\newblock In {\em NeurIPS}, 1999.

\bibitem{Hu_2019_CVPR}
Di Hu, Feiping Nie, and Xuelong Li.
\newblock Deep multimodal clustering for unsupervised audiovisual learning.
\newblock In {\em Proc. CVPR}, June 2019.

\bibitem{izadinia2012multimodal}
Hamid Izadinia, Imran Saleemi, and Mubarak Shah.
\newblock Multimodal analysis for identification and segmentation of
  moving-sounding objects.
\newblock {\em IEEE Trans. Multimed.}, 2012.

\bibitem{khosravan2018attention}
Naji Khosravan, Shervin Ardeshir, and Rohit Puri.
\newblock On attention modules for audio-visual synchronization.
\newblock In {\em Proc. CVPR Workshop}, 2019.

\bibitem{kidron2005pixels}
Einat Kidron, Yoav~Y Schechner, and Michael Elad.
\newblock Pixels that sound.
\newblock In {\em Proc. CVPR}, 2005.

\bibitem{Lin2017}
Tsung-Yi Lin, Priya Goyal, Ross Girshick, Kaiming He, and Piotr Dollár.
\newblock Focal loss for dense object detection.
\newblock In {\em Proc. ICCV}, 2017.

\bibitem{Marcheret15}
Etienne Marcheret, Gerasimos Potamianos, Josef Vopicka, and Vaibhava Goel.
\newblock Detecting audio-visual synchrony using deep neural networks.
\newblock In {\em Proc. ICSA}, 2015.

\bibitem{oord2018representation}
Aaron van~den Oord, Yazhe Li, and Oriol Vinyals.
\newblock Representation learning with contrastive predictive coding.
\newblock {\em arXiv preprint arXiv:1807.03748}, 2018.

\bibitem{owens18audio-visual}
Andrew Owens and Alexei~A. Efros.
\newblock Audio-visual scene analysis with self-supervised multisensory
  features.
\newblock In {\em Proc. {ECCV}}, 2018.

\bibitem{Owens2018b}
Andrew Owens and Alexei~A. Efros.
\newblock Audio-visual scene analysis with self-supervised multisensory
  features.
\newblock In {\em Proc. ECCV}, 2018.

\bibitem{qian2020multiple}
Rui Qian, Di Hu, Heinrich Dinkel, Mengyue Wu, Ning Xu, and Weiyao Lin.
\newblock Multiple sound sources localization from coarse to fine.
\newblock In {\em Proc. ECCV}, 2020.

\bibitem{Ramaswamy_2020_WACV}
Janani Ramaswamy and Sukhendu Das.
\newblock See the sound, hear the pixels.
\newblock In {\em Proc. WACV}, 2020.

\bibitem{Ren16}
Shaoqing Ren, Kaiming He, Ross Girshick, and Jian Sun.
\newblock Faster {R-CNN}: Towards real-time object detection with region
  proposal networks.
\newblock In {\em NeurIPS}, 2016.

\bibitem{rouditchenko2019self}
Andrew Rouditchenko, Hang Zhao, Chuang Gan, Josh McDermott, and Antonio
  Torralba.
\newblock Self-supervised audio-visual co-segmentation.
\newblock In {\em Proc. ICASSP}, 2019.

\bibitem{senocak2018learning}
Arda Senocak, Tae-Hyun Oh, Junsik Kim, Ming-Hsuan Yang, and In~So Kweon.
\newblock Learning to localize sound source in visual scenes.
\newblock In {\em Proc. CVPR}, 2018.

\bibitem{shrivastava2016training}
Abhinav Shrivastava, Abhinav Gupta, and Ross Girshick.
\newblock Training region-based object detectors with online hard example
  mining.
\newblock In {\em Proc. CVPR}, 2016.

\bibitem{tao2018scale}
Xin Tao, Hongyun Gao, Xiaoyong Shen, Jue Wang, and Jiaya Jia.
\newblock Scale-recurrent network for deep image deblurring.
\newblock In {\em Proc. CVPR}, 2018.

\bibitem{Thomee16}
Bart Thomee, David~A Shamma, Gerald Friedland, Benjamin Elizalde, Karl Ni,
  Douglas Poland, Damian Borth, and Li-Jia Li.
\newblock Yfcc100m: the new data in multimedia research.
\newblock {\em Commun. ACM}, 2016.

\bibitem{tian2018audio}
Yapeng Tian, Jing Shi, Bochen Li, Zhiyao Duan, and Chenliang Xu.
\newblock Audio-visual event localization in unconstrained videos.
\newblock In {\em Proc. ECCV}, 2018.

\bibitem{viola01brobust}
Paul Viola and Michael Jones.
\newblock Robust real-time object detection.
\newblock In {\em Proc. SCTV Workshop}, 2001.

\bibitem{tian2020avvp}
Dingzeyu~Li Yapeng~Tian and Chenliang Xu.
\newblock Unified multisensory perception: Weakly-supervised audio-visual video
  parsing.
\newblock In {\em Proc. ECCV}, 2020.

\bibitem{zhao2019sound}
Hang Zhao, Chuang Gan, Wei-Chiu Ma, and Antonio Torralba.
\newblock The sound of motions.
\newblock In {\em Proc. ICCV}, 2019.

\bibitem{zhao2018sound}
Hang Zhao, Chuang Gan, Andrew Rouditchenko, Carl Vondrick, Josh McDermott, and
  Antonio Torralba.
\newblock The sound of pixels.
\newblock In {\em Proc. ECCV}, 2018.

\end{thebibliography}
